\documentclass{article}

\PassOptionsToPackage{numbers}{natbib}
\usepackage[final]{neurips_2022}
\usepackage{wrapfig}

\usepackage[utf8]{inputenc} 
\usepackage[T1]{fontenc}    

\usepackage{url}            
\usepackage{booktabs}       
\usepackage{amsfonts}       
\usepackage{nicefrac}       
\usepackage{microtype}      

\usepackage{tikz}
\usepackage{amsmath}
\usepackage{amssymb}
\usepackage{mathtools}
\usepackage{amsthm}
\usepackage{pgf}

\usepackage{placeins}
\usepackage{mathtools}
\usepackage[compact]{titlesec}
\titlespacing*{\paragraph}{0pt}{0\baselineskip}{0.5em}
\titlespacing{\section}{0pt}{0.5ex}{0.5ex}
\titlespacing{\subsection}{0pt}{0.5ex}{0ex}
\titlespacing{\subsubsection}{0pt}{0ex}{0ex}
\theoremstyle{plain}
\newtheorem{theorem}{Theorem}[section]

\newtheorem{lemma}[theorem]{Lemma}

\newtheorem{definition}[theorem]{Definition}

\theoremstyle{definition}

\theoremstyle{remark}

\usepackage{optidef}
\usepackage{centernot}
\usepackage{verbatim}
\usepackage{caption}
\usepackage{subcaption}

\usepackage{amsthm,thmtools}%
\theoremstyle{definition}

\theoremstyle{plain}

\DeclareMathOperator{\X}{\mathcal{X}}

\DeclareMathOperator{\Z}{\mathcal{Z}}
\newcommand{\setfilter}[2]{\{ #1 \mid #2\}}

\usepackage[symbol]{footmisc}

\usepackage{hyperref}       
\usepackage{xcolor}         
\usepackage[capitalize]{cleveref}

\title{Zonotope Domains for Lagrangian Neural Network Verification}

\author{%
  Matt Jordan\thanks{Equal contribution}\\
  UT Austin \\
  \texttt{mjordan@cs.utexas.edu} \\
  \And 
  Jonathan Hayase\footnotemark[1]\\
  University of Washington \\
  \texttt{jhayase@cs.washington.edu}\\
  \And 
  Alexandros G. Dimakis\\
  UT Austin \\
  \texttt{dimakis@austin.utexas.edu}\\
  \And 
  Sewoong Oh\\
  University of Washington\\
  \texttt{sewoong@cs.washington.edu}\\
  \\
}

\begin{document}

\maketitle

\begin{abstract}
\footnotetext[2]{Github Repo: \url{https://github.com/revbucket/dual-verification}}

Neural network verification aims to provide provable bounds for the output of a neural network for a given input range. Notable prior works in this domain have either generated bounds using abstract domains, which preserve some dependency between intermediate neurons in the network; or framed verification as an optimization problem and solved a relaxation using Lagrangian methods. A key drawback of the latter technique is that each neuron is treated independently, thereby ignoring important neuron interactions. We provide an approach that merges these two threads and uses zonotopes within a Lagrangian decomposition. Crucially, we can decompose the problem of verifying a deep neural network into the verification of many 2-layer neural networks. While each of these problems is provably hard, we provide efficient relaxation methods that are amenable to efficient dual ascent procedures. Our technique yields bounds that improve upon both linear programming and Lagrangian-based verification techniques in both time and bound tightness.
\end{abstract}

\section{Introduction}
With the growing prevalence of machine learning in real-world applications, 
the brittleness of deep learning systems poses an even greater threat. 
It is well-known that deep neural networks are vulnerable to adversarial examples, where a minor change in the input to a network can cause a major change in the output \cite{Szegedy2013-yt}. 
There is a long history of defense techniques being proposed to improve the robustness of a network, only to be completely broken shortly thereafter \cite{Athalye2018-vt}. 
This has inspired researchers to focus instead on
{\em neural network verification}, which, for a given network and a range of input, aims to verify whether a certain property is satisfied for every input in that range.
A typical adversarial attack seeks to minimize the output of a scalar-valued network subject to certain input constraints. Verification provides lower bounds on the minimum output value achievable by such an adversary. 
Concretely, for a scalar-valued network $f$ and input range $\mathcal{X}$, verification provides lower bounds to the problem 
\( \min_{x\in\mathcal{X}} f(x)\).

As observed in \cite{Salman2019-jh}, prior works in verification can be broadly categorized into primal and dual views. In the primal view, convex relaxations are applied to attain, for every intermediate layer of the network, a convex superset of the true attainable range. For example, if $f=f_L\circ \cdots \circ f_1$, convex sets $\Z_k$ are obtained such that $\setfilter{f_k\circ \cdots \circ f_1(x)}{x\in \X} \subseteq \Z_k$ for $k\in\{1,\ldots,L\}$.

Under the dual lens, verification is treated as a stagewise optimization problem and Lagrangian relaxation is applied to yield a dual function that always provides valid lower bounds. Often, this dual function is decomposable into a sum of minimization problems which are efficiently computable. One hallmark of all existing dual verification techniques is that intermediate bounds $\Z_k$ are required. In this case, either an efficient primal verification algorithm must first be applied or, for example, the dual verifier can be run iteratively on each neuron to provide upper and lower bounds for each intermediate layer.

Our approach is an attempt to combine the primal and dual threads of verification. We first apply a primal verification algorithm that generates bounds on the attainable range of each intermediate layer.
To do this we leverage zonotopes, a more expressive class of polytopes than axis-aligned hyperboxes.
Notably, we offer an improvement to existing zonotope bounds when we are also provided with incomparable hyperbox bounds.
These zonotopic intermediate bounds are then applied to a dual verification framework, for which dual ascent can be performed.
Our formulation may be viewed as taking the original nonconvex verification problem and decomposing it into many subproblems, where each subproblem is a verification problem for a 2-layer neural network.
However, as even verifying a 2-layer network is hard in general, we further develop efficient relaxation techniques that allow for tractable dual ascent.


We introduce a novel algorithm, which we call ZonoDual, which \textit{i}) is highly scalable and amenable to GPU acceleration, \textit{ii}) provides tighter bounds than both the prior primal and dual techniques upon which our approach is built, \textit{iii}) is highly tunable and able to effectively balance the competing objectives of bound tightness and computation speed, and \textit{iv}) is applicable as an add-on to existing dual verification frameworks to further boost their performance. We apply ZonoDual to a variety of networks trained on MNIST and CIFAR-10 and demonstrate that ZonoDual outperforms the linear programming relaxation in both tightness and runtime, and yields a tighter bounding algorithm than the prior dual approaches.



We first discuss  prior works in the verification domain. 
Then, we examine the existing dual framework that serves as the backbone for our algorithm, paying particular attention to the areas in which this may be tightened. 
Then, we discuss how we attain zonotope-based intermediate bounds and introduce the fundamental sub-problem required by our dual ascent algorithm. 
Ultimately, we combine these components into our final   algorithm 
and demonstrate the scalability and tightness of our approach on networks trained on the MNIST and CIFAR-10 datasets.
\section{Related Work}
Neural network verification is a well-studied problem and has been approached from several angles. Exactly solving the verification problem is known as \emph{complete verification}, and is known to be NP-hard even for 2-layer ReLU networks \cite{Katz2017-qz}. Complete   verification approaches  include  mixed-integer programs, satisfiability modulo theories (SMT), geometric procedures, and branch-and-bound techniques \cite{Tjeng2017-qp, Fischetti2018-mi, anderson2020strong, Katz2017-qz,Jordan2019-fv, De_Palma2021-sp, Xu2020-du}. In particular, branch-and-bound procedures generate custom branching rules to decompose verification into many smaller subproblems, which can be relaxed or further branched, allowing the algorithm to focus its efforts on the areas for greatest improvement \cite{lu2019neural, De_Palma_undated-ie}. We stress that this current work focuses only on the `bound' phase, and our approach may be applied to branched subproblems. 

Often it is not necessary to exactly solve the verification problem, and only providing a lower bound to the minimization problem can suffice. This is known as \emph{incomplete verification}. We closely examine the primal and dual verification threads, but first mention a few lines of  work that do not fit neatly into this paradigm: notable approaches here include semidefinite programming relaxations \cite{Raghunathan2018-zu, dathathri2020enabling}, or verification via provable upper bounds of the Lipschitz constant \cite{Fazlyab2019-im, Virmaux2018-ti, Hashemi2020-al, Jordan2020-tq}. 

\paragraph{Primal verification:}
A general theme in primal verification techniques is to generate convex sets that bound the attainable range of each intermediate layer. The simplest such approach is interval bound propagation (IBP) which computes a bounding hyperbox for each layer \cite{Ehlers2017-ou}. A more complex approach is to bound each intermediate layer with a polytope. Several different polyhedral relaxations are available, with DeepPoly being on the lesser-complexity side, PLANET being the canonical triangle relaxation, and the formulation by Anderson et. al being tightest albeit with exponential complexity \cite{Zico_Kolter2017-va, SinghGagandeep2019-ki, anderson2020strong}. Zonotopes provide a happy medium, being both highly expressive and computationally easy to work with \cite{Singh2018-tg, singh2018boosting}. We will discuss zonotope relaxations more thoroughly in section 4.

\paragraph{Dual verification:}
Orthogonal to the primal approaches exist several works which consider verification by relaxations in a dual space \cite{Krishnamurthy2018-ow, Bunel2020-cl, De_Palma_undated-ie, Berrada2021-tc, chen2021deepsplit}. Here, verification is viewed as an optimization problem with constraints enforced by layers of the network. The general scheme here is to introduce dual variables penalizing constraint violation, and then leverage weak duality to provide valid lower bounds. The standard Lagrangian relaxation was first proposed in \cite{Krishnamurthy2018-ow}, and then several improvements have been made to provide provably tighter bounds, faster convergence, or generalizations of the augmented lagrangian. These techniques are highly scalable and notably, one recent work has been able to provide bounds tighter than even the standard LP relaxation \cite{De_Palma2021-sp}. Our approach will follow a similar scheme, but differs fundamentally from prior works in that we operate over a tighter primal domain and can therefore provide tighter bounds. 
\section{Lagrangian Decomposition}


\paragraph{Problem Definition:}
In a canonical neural network verification problem, we are given a \emph{scalar-valued} feedforward neural network, $f(\cdot)$, composed of $L$ alternating compositions of affine layers and elementwise ReLU layers. We name the intermediate representations as $z_0\dots z_L$, and are supplied with a subset, $\X$, of the domain of $f(\cdot)$, which we assume to be an axis-aligned hyperbox. Verification is then written as an optimization problem 
\begin{subequations}
\label{eq:optim}
\begin{alignat}{3}
\underset{x_0\in\X,z_0,\ldots,z_L}{\mathrm{minimize}}
& & z_L\label{eq:og-opt}\\
\mathrm{subject\:to}\hspace{0.4em}
&& z_0 &= W_0x_0\\
&& z_{k+1} &= W_{k+1}\sigma(z_k) &\quad& (0\leq k \leq L-1)
\end{alignat}
\end{subequations}
where $\sigma$ denotes the ReLU operator applied elementwise, and each $W_k$ is the weight of an affine or convolutional layer. Our formulation is also amenable to bias terms everywhere, though we omit these throughout for simplicity.
%
\paragraph{Dual verification:}
While there are several verification approaches leveraging Lagrangian duality, we consider the Lagrangian decomposition formulation introduced in \cite{Bunel2020-cl}. Here, each primal variable $z_k$ of the verification problem is replaced by a pair of primal variables $z_k^A, z_k^B$, along with accompanying equality constraints, yielding an equivalent optimization problem:
\begin{subequations}
\begin{alignat}{3}
\underset{x_0\in\X,z_0^A,\ldots,z_{L-1}^B,z_L^A}{\mathrm{minimize}}
& & z_L^A\label{eq:og-decomp-opt}\\
\mathrm{subject\:to}\hspace{1.6em}
&& z_0^A &= W_0x_0\\
&& z_{k+1}^A &= W_{k+1}\sigma(z_k^B) &\quad& (0\leq k \leq L-1)\\
&& z_{k}^A &= z_{k}^B && (0 \leq k \leq L-1) \label{eq:og-decomp-opt-eq}
\end{alignat}
\end{subequations}
By introducing unconstrained dual variables $\rho_k$ for each constraint in Equation \eqref{eq:og-decomp-opt-eq} and maximizing over $\rho_k$, we attain an optimization problem that is equivalent to the original. To tractably solve this, two relaxations are performed. First, weak duality is applied by swapping the max and min, which introduces the dual function $g(\rho)$, for which every value of $\rho$ provides a lower bound: 
\begin{subequations}
\begin{alignat}{3}
g(\rho) \coloneqq \underset{x_0\in\X,z_0^A,\ldots,z_{L-1}^B, z_L^A}{\mathrm{minimize}}
& & \mathclap{\hspace{6em}z_L^A + \sum_{k=0}^{L-1}\rho_k^\top (z_k^A-z_k^B)}\label{eq:og-decomp-dual}\\
\mathrm{subject\:to}\hspace{0.4em}
&\hspace{0.5em}& z_0^A &= W_0x_0\\
&& z_{k+1}^A &= W_{k+1}\sigma(z_k^B) &\quad& (0\leq k \leq L-1),
\end{alignat}
\end{subequations}
The goal now is to solve $\max_\rho g(\rho)$. The dual function can be decomposed into $L$ subproblems by substituting equality constraints and rearranging the objective function:
\begin{subequations}
\begin{align}
    g(\rho) \coloneqq &\left(\min_{x_0\in\X} \rho_0^\top W_0 x_0 \right) +  \sum_{k=1}^{L-2} \left(\min_{z_k^B} \rho_{k+1}^\top W_{k+1}\sigma(z_k^B) - \rho_k^\top z_k^B\right) \label{eq:dual1} +\\
               &\left(\min_{z_{L-1}^B} W_L\sigma(z_{L-1}^B) - \rho_{L-1}^\top z_{L-1}^B\right) \label{eq:dual2}
\end{align}
\end{subequations}
In its current state, it is unlikely that $g(\rho)$ for nonzero $\rho$ will lead to any finite lower bound as each $z_k^B$ is unconstrained. In other words, only subproblem, \eqref{eq:dual1}, has any information about the input constraint $(x_0\in\X)$. To remedy this, one can tighten the dual optimization by imposing \emph{intermediate bounds} on $z_k^B$'s, so long as those bounds include all feasible values.
Specifically, for each $z_k^B$, we can impose a bounding set $\Z_k$, as long as the following implication holds:
\begin{align}
    x_0\in \X \implies z_k^B \in \Z_k &\quad\quad 1\leq k \leq L-1. \label{eq:intermed-bounds}
\end{align}

All prior dual approaches have chosen the sets $\Z_k$ to be axis-aligned hyperboxes for two reasons. First, it is extremely efficient to attain hyperbox bounds at every layer. Second, when $\Z_k$ is a hyperbox, the dual function can be evaluated efficiently by further decomposing each component of the dual along its coordinates. Equipped with intermediate bounds $\Z_k$, and noting that $g(\rho)$ is a concave function of $\rho$, the standard procedure is to perform dual ascent on $g(\rho)$. This requires gradients $\nabla_\rho g(\rho)$, which can easily be computed as the primal residuals. Letting $z_k^{A*}$ and $z_k^{B*}$ be the argmin of the dual function, then $\nabla_{\rho_k}g(\rho)= z_k^{A*} - z_k^{B*}$.

The key innovation in this work is to replace the intermediate bounds $\Z_k$ with zonotopes everywhere. In this case, intermediate bounds may easily be computed as we will see in the next section. Zonotopes have a distinct advantage over hyperboxes in that every coordinate is no longer independent, and therefore neuron dependencies are encoded. This further constrains the feasible set of the dual function and leads to larger dual values. However, this comes with the cost that the dual function is more computationally difficult to evaluate.  


\section{Zonotopes}
\label{sec:zono}

We start with a review of zonotopes and how they may be used to attain intermediate bounds. Then we introduce the problem of ReLU programming, tie it to the dual decomposition formulation, and discuss our relaxations for efficient dual evaluation. Proofs of all claims are contained in the appendix.

\paragraph{Zonotope Properties:}
Zonotopes are a class of polytopes, which we formally define using the notation $Z(c,E)$ to refer to the set
\[Z(c, E) := \setfilter{c+Ey}{y\in [-1,1]^m}\]
where $c\in \mathbb{R}^d$ is called the \emph{center}, and $E\in \mathbb{R}^{d\times m}$ is called the \emph{generator matrix}. Each column of the generator matrix is a line segment in $\mathbb{R}^d$, and $Z(c,E)$ can be equivalently be viewed as the Minkowski sum of each generator column offset by $c$; or as the affine map $y \to c+Ey$ applied to the $\ell_\infty$ ball in $\mathbb{R}^m$. In particular, note that a  hyperbox is a zonotope with a diagonal generator matrix. Zonotopes have several convenient properties:
 
 \emph{Efficient Linear Programs:} Linear programs can be solved in closed form over a zonotope:
    \[
        \min_{z\in Z(c,E)} a^\top z = a^\top c + |a^\top E|\vec{1}.\\
  \]
\emph{Closure under affine operators and Minkowski sums:} given an affine map $x\to Wx+b$, the set $\setfilter{Wz+b}{z\in Z(c,E)}$ is equivalent to the zonotope $Z(Wc+b, WE)$. The Minkowski sum of two sets $A, B$ is defined as $A\oplus B := \setfilter{x +y}{x\in A,\quad y\in B}$. Letting  $E_1||E_2$ denotes concatenation of columns, the Minkowski sum of two zonotopes $Z(c_1, E_1)$, $Z(c_2, E_2)$ is also a zonotope:
    \[Z(c_1, E_1) \oplus Z(c_2, E_2) = Z(c1 + c2, E_1 || E_2).\]
\paragraph{Intermediate Bounds with Zonotopes:} Prior work has developed efficient techniques to generate intermediate layer bounds using zonotopes for feedforward neural networks. The details are deferred to the appendix, and we can assume that we have access to a black box that, given a zonotope $\Z$, is able to generate a zonotope $\mathcal{Y}$ such that $\setfilter{\sigma(z)}{z\in \Z} \subseteq \mathcal{Y}$. Such an operation is called a \emph{sound pushforward operator}. One such pushforward operator is known as \texttt{DeepZ} and can be applied repeatedly to provide valid zonotopic bounds for every intermediate layer of the neural network \cite{Singh2018-tg}. We make two observations regarding these intermediate bounds here: 
\begin{restatable}{prop}{kwProp}\label{prop:zono-kw}
Let $\Z_k$ be the $k$\textsuperscript{th} zonotope bound provided by the \texttt{DeepZ} procedure, and let $\mathcal{H}_k$ be the $k$\textsuperscript{th} intermediate bound provided by the Kolter-Wong dual procedure \cite{Zico_Kolter2017-va}. 
Then the smallest hyperbox containing $\Z_k$ is exactly $\mathcal{H}_k$.
\end{restatable}

\begin{restatable}{prop}{hboxProp}\label{prop:box-deepz}
Suppose $\Z$ is a zonotope and $\mathcal{H}$ is a hyperbox, such that $\Z \centernot \subseteq \mathcal{H}$. Then we can develop a sound pushforward operator for the ReLU operator that outputs a zonotope $\Z'$ containing the ReLU of every element of $\Z\cap \mathcal{H}$, with the property that $\Z' \subsetneq \texttt{DeepZ}(\Z)$.
\end{restatable}
The first proposition states that a common technique for generating intermediate hyperbox bounds is never any tighter than the procedure we use to generate intermediate zonotope bounds. 
The second proposition describes an improved pushforward operator that offers improvements against $\texttt{DeepZ}$ but requires extra information provided by a hyperbox.

\paragraph{ReLU Programming:}
We return our attention to the dual function introduced in equations \Cref{eq:dual1,eq:dual2}. Note that the first term of Equation \ref{eq:dual1} is a linear program of a zonotope which can be solved extremely efficiently, while all other terms adopt a form we refer to as a \emph{ReLU program}
\begin{definition}
Given a set $\Z \subseteq \mathbb{R}^d$, and two objective vectors $c_1, c_2 \in \mathbb{R}^d$, we say the \textbf{ReLU Program} is the optimization problem
\begin{equation}\label{eq:relu-program}
    \min_{z\in\Z} c_1^\top z + c_2^\top \sigma(z).
\end{equation}
\end{definition}
ReLU programming is nonconvex for general $c_2$. When $\Z$ is an interval, only the endpoints, or 0 if it is contained in the interval, are candidate optima. In this case, the ReLU program may be solved by evaluating the objective at each candidate. When $\Z$ is a hyperbox in $\mathbb{R}^d$, each coordinate can be considered independently, amounting to solving $d$ ReLU programs over intervals. The story quickly becomes more complicated if $\Z$ is a zonotope:
\begin{restatable}{thm}{rphardness}
When $\Z$ is a zonotope, solving a ReLU program over $\Z$ is equivalent to a neural network verification problem for a 2-layer network, which is known to be NP-hard.
\end{restatable}

\textbf{Remarks: } This theorem implies that solving ReLU programs over more complex sets, such as polytopes in general, is also NP-hard. However it remains an open question whether solving ReLU programs over other simpler sets such as parallepipeds, i.e. zonotopes with an invertible generator matrix, is hard. Note that ReLU programs can be solved exactly via a Mixed-Integer Program (MIP) using the standard form for encoding a ReLU \cite{Tjeng2017-qp}.

We pause here to observe the current situation. Lagrangian methods for verification first make a relaxation by applying weak duality. When using zonotopes for the intermediate bounds, one is able to decompose the very large non-convex optimization problem of deep network verification into many smaller subproblems which are equivalent to two-layer network verification problems. Unfortunately, each of these subproblems is also NP-hard, and thus must be further relaxed to attain tractability. 

\paragraph{Zonotope Partitioning:}
The primary relaxation technique we consider comes from the following observation. Consider a $d$-dimensional zonotope $\Z$ and ReLU programming objectives $(c_1, c_2)$. 
Then, for any partition \(S_1 \cup \cdots \cup S_n\) of $\{1,\dots, d\}$, the inequality
\begin{align*}
    \MoveEqLeft{\min_{z\in\Z} \sum_{i=1}^d \big(c_{1,i}z_i + c_{2,i}\sigma(z_i)\big)}
    \geq \sum_{j=1}^n\min_{z\in\Z} \sum_{i\in S_j} \big(c_{1,i}z_i + c_{2,i}\sigma(z_i)\big) 
\end{align*}
holds. This approach has a nice geometric interpretation as well.
The decomposed minimization $\min_{z\in \Z} \sum_{i\in S_j} \big(c_{1,i}z_i + c_{2,i}\sigma(z_i) \big)$ can be viewed as optimization over $\Z$ when it has been projected onto the linear subspace spanned by the elements of $S_j$. Projection onto a linear subspace is an affine operator, so each projection of $\Z$ is itself a zonotope. 
Indeed, the projection onto the linear subspace spanned by $S_j$ is simply the center and generator rows indexed by $S_j$.

Observe that if each $S_j$ were a singleton set, then decomposing the ReLU program according to these $S_j$'s is equivalent to relaxing the ReLU program by casting $\Z$ to the smallest containing hyperbox. If the partition were the trivial partition $\{1,\dots, d\}$, then $\Z$ remains unchanged. Finally, notice that the Minkowski sum of every projection of $\Z$ is itself a zonotope. 

In this sense, by choosing any partitioning of coordinates, we are able to relax $\Z$ into a zonotope that is simultaneously a superset of $\Z$ and a subset of the hyperbox-relaxation of $\Z$. This is optimistic for our dual ascent procedure, as any partitioning restricts the primal feasible space for the dual function evaluation. It is in this way that we are able to interpolate the coarseness of our relaxation, which is directly controlled by the coarseness of our partitioning. Because we expect the complexity of solving a ReLU program over a zonotope to be superlinear in the dimension, shattering a zonotope in this fashion can drastically improve the complexity of solving a relaxation of a ReLU program.

\paragraph{Two-dimensional Zonotopes:}
A special case we consider is when we partition a zonotope into 2-dimensional groups. This is because a 2-dimensional zonotope can be significantly more descriptive than a 2-dimensional rectangle, but the solution of a ReLU program can still be computed efficiently. Observe that the argmin of a ReLU program over a 2-dimensional zonotope must occur at either i) a vertex of the zonotope, ii) a point where the zonotope crosses a coordinate-axis, iii) the origin. This follows from the fact that a ReLU program over 2-dimensions may be viewed as 4 independent linear programs over the intersection of the zonotope and each orthant. 
For the purposes of ReLU programs, it suffices to enumerate all vertices of the zonotope and compute any axis crossings or origin-containments. It turns out that this may be done in nearly linear time:
\begin{restatable}{thm}{zonoTwoD}\label{thm:2dzono-vertices}
If $Z(c,E)$ is a 2-dimensional zonotope with $m$ generators, $Z$ has $2m$ vertices and the set of all vertices, axis crossings, and the containment of the origin can all be computed in $O(m\log m)$ time.
\end{restatable}
\section{The ZonoDual Algorithm}

With both preliminaries of dual verification techniques and zonotopic primal verification techniques in hand, we can now fully describe the algorithm we employ in practice. We first describe the algorithm as it is used to bound a single neuron's value, i.e. the output neuron. After we describe how we adapt our approach to the stagewise setting where our algorithm is applied to each intermediate neuron to yield tighter box bounds. Our algorithm can be described in three main phases: an \emph{initialization phase}, where primal bounds are generated to define the feasible set for each subproblem in the dual function; an \emph{iteration phase}, where the argmin of the dual function is computed many times to provide informative gradients for updates to the dual variables; and an \emph{evaluation phase}, where the primal bounds are tightened to yield a more computationally intensive dual function, which only needs to be evaluated a single time. 

\paragraph{Initialization phase:} Naively, we can apply the \texttt{DeepZ} procedure here to attain a zonotope bounding the output range of the network at every layer. Both this procedure and our box-improved variant can be done extremely efficiently. We adapt two techniques from \cite{Bunel2020-cl} in the initialization phase. First we notice that some very minor improvements to \texttt{DeepZ} can be made using an application of Proposition \ref{prop:box-deepz} when provided with the boxes from interval bound propagation. Second, we initialize the dual variables to the KW dual variables from \cite{Zico_Kolter2017-va}, which by \ref{prop:zono-kw} are equivalent to the scale factors applied internally in the zonotope pushforward operators.

\paragraph{Iteration Phase:} In the iteration phase, each intermediate zonotope $\Z_k$ is partitioned into components that are amenable to repeated evaluations of the dual function. We always initially partition each zonotope into 2-d chunks for which the dual function may be evaluated on a GPU without requiring any MIP calls. Recall that for a $d$-dimensional zonotope with $m$ generators, after the initial $O(dm\log m)$ candidate optima computation, each ReLU program can be evaluated in $O(dm)$ time. These initial 2-d partitions are computed using one of several simple heuristics, described in the appendix, with the aim to generate 2-d zonotopes which are as `un-box-like' as possible. Optionally, after an initial 2-d iteration phase, these partitions may be combined and dual ascent can be applied, where each subproblem now requires several calls to a MIP solver. At any point during this phase, the dual function evaluations still provide valid lower bounds to the verification problem. 

\paragraph{Evaluation Phase:} In this phase, we can presume that effective dual variables have been found during the previous phase. The objective here is to tighten the feasible range of each subproblem, which will necessarily improve the value of the dual function. Because we only need to evaluate a bound on the dual function once here, we find that it is often worth it to spend more time solving each subproblem. We do this by simply merging several partition elements together, converting our 2-dimensional zonotopes into higher-dimensional ones that require calls to a MIP solver to compute. Despite the theoretical intractibility of MIPs, we find that this procedure is remarkably efficient. In practice, it is often sufficient to only merge zonotopes at the later layers in a network where the dual function subproblems are the most negative. Finally, we note that as only a \emph{bound} upon the dual function is required here, and each subproblem is itself a 2-layer verification problem, the MIPs may be terminated early. Any other incomplete verification approach can be applied here as well.

\paragraph{Guarantees:} Ultimately, our algorithm will always provide a valid lower bound to the verification problem so long as the dual function is evaluated appropriately. The provable correctness of our algorithm stems from i) the fact that weak duality always holds and any evaluation of the dual function with feasible dual variables provides a valid lower bound tot the primal problem; and ii) by leveraging zonotopes and a sound relaxation of them, we only ever provide lower bounds to the evaluation of the dual function. Further, we can also guarantee that, for any fixed set of dual variables $\rho$, the bounds we provide will be tighter than those provided by the approach in \cite{Bunel2020-cl}. This follows because we have reduced the feasible set within each dual subproblem, relative to the hyperbox approaches in prior works. Similarly, because $\rho=0$ is always a valid choice, when the dual variables are optimized appropriately, we will provide tighter bounds than whichever primal verification method is applied at the initialization phase. 

\paragraph{Stagewise computation:} When the network of interest is of sufficiently small size, any verification algorithm may be applied stagewise, where the algorithm computes upper and lower bounds to each neuron, starting from those in the first layer. This is often effective for generating tighter intermediate box bounds, which in turn yields much tighter output bounds. However this comes the cost of running many calls to the verification algorithm. This can be parallellized for verification procedures amenable to GPU acceleration, but demands a large memory footprint.

At first glance, it may seem that such a stagewise procedure obviates the improvements we attain by leveraging zonotopes: indeed, each stagewise bound is a \emph{box bound}. However, we show that we can apply this stagewise trick to our approach to generate tighter box bounds, which can then be employed by Proposition \ref{prop:box-deepz} to yield tighter zonotopic bounds. Additionally, in the case that these box bounds are incomparable to the improved zonotope bounds, these can be employed in imposing further constraints upon each subproblem. Instead of solving each ReLU program over a zonotope, we can instead solve each ReLU program over the intersection of a zonotope and a hyperbox. In the higher-dimensional case, it is trivial to apply elementwise constraints to each variable in the MIP --- in fact, this often improves the running time of each MIP. In the 2-d case, we observe that we are able to solve ReLU programs over the intersection of a 2-d zonotope and rectangle as efficiently as solving over a 2-d zonotope alone. This is the content of the following corollary to Theorem \ref{thm:2dzono-vertices}. 
\begin{restatable}{cor}{zonoTwoDBox}\label{cor:2d-zono-box}
Given a 2-d zonotope with $m$ generators, the set of candidate optimal points to a ReLU program has cardinality no greater than $(2m+9)$ and is computable in time $O(m\log m)$.
\end{restatable}
\paragraph{Ablation study:} To clarify the bound improvements and their corresponding runtime cost imposed by each step of our approach, we perform an ablation study on an MNIST fully connected network (MNIST FFNet). We report the bounds relative to the bound attained by the LP relaxation in Table \ref{table:ablation-mini}. We consider three initialization settings separately: the standard \texttt{DeepZ} zonotopes, the \texttt{DeepZ} zonotopes augmented with IBP bounds, and zonotopes informed by the the box bounds from BDD+ \cite{Bunel2020-cl}. Then we perform dual ascent over 2-d zonotopes, before merging the zonotopes of the final layer only and evaluating the MIP. We report these numbers in Table \ref{table:ablation-mini}. Here we see that a tight initialization does indeed help, but this gap shrinks after our dual ascent procedure. We also see a substantial gain in bound improvement by using MIPs, even in the last layer's zonotopes alone.

\begin{table}[h]
\vspace*{-.4cm}
\centering
\caption{Ablation study. We report the bound obtained and cumulative runtime at four points during an extended run of our algorithm: 1) 1000 iterations of 2D dual ascent; 2) Evaluating the dual with a medium partitioning size; 3) 50 steps of dual ascent using MIPs; 4) Evaluating the dual with a coarser partitioning. We report results when zonotope preactivations are used and when they are combined with the hyperbox preactivations of BDD+.}\label{table:ablation-mini}
\begin{tabular}{l|rrrr} 
\toprule
Bounds\textbackslash{}Phase & 1      & 2      & 3      & 4     \\
\midrule
Zonotopes & -16.52 & -11.73 & -10.95 & -5.63\\
Runtime (s) & 6.0 & 7.4 & 161 & 169 \\[0.5ex]
\midrule
\(\text{Zono} \cap \text{BDD+}\) & -13.02 & -9.33  & -8.96  & -4.47 \\
Runtime (s) & 11.3 & 12.5 & 101 & 107\\ 
\bottomrule
\end{tabular}
\end{table}

\section{Experiments}

We evaluate the effectiveness of our approach on several networks trained on the MNIST and CIFAR-10 datasets. While many recent verification techniques employ a branch-and-bound strategy, we only provide innovations on the bounding component of this pipeline. Strong verifiers should employ both effective bounding algorithms and clever branching strategies, which are both necessary but fairly independent components. Like other dual-based verification techniques, our approach can be applied using any of a number of branching strategies. We focus our efforts primarily on the improvements in the \emph{bound} provided by a single run of each considered method. We argue this is a reasonable choice of metric because most prior benchmarks focus on $\epsilon$-values far lower than typically used in adversarial attack scenarios. This artifically decreases the depth to which branching strategies must search, skewing the results. By focusing on the bound of the verification, we directly measure the metric that bounding algorithms such as ours seek to optimize. However, for the sake of completeness, we do also evaluate verified robustness percentage upon several benchmark MNIST networks.
\begin{figure}[!htb]
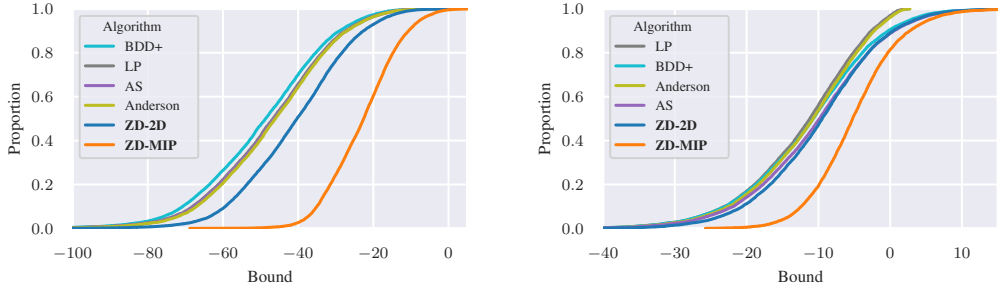

\vspace{-1ex}
\begin{subfigure}{0.5\textwidth}
  \centering
    \resizebox{0.9\textwidth}{!}{\input{figures/mnist_deep_cdf.pgf}}
\end{subfigure}
\begin{subfigure}{0.5\textwidth}
  \centering
    \resizebox{0.9\textwidth}{!}{\input{figures/cifar_sgd_cdf.pgf}}
\end{subfigure}
\vspace{-1ex}
    \caption{Cumulative density function plots of the bounds obtained by each algorithm for a deep convolutional network trained on MNIST (left) and a deep convolutional network trained on CIFAR10 (right). Lines further to the right indicate tighter bounds. Our algorithms are bolded in the legend.} 
    \label{fig:single-stage}
\end{figure}
For bound evaluations, we separate our experiments into two cases: we first examine the setting where only the output neuron is optimized, which we denote the `single-stage' setting; and the case where every neuron is optimized to provide tighter intermediate box bounds, denoted the `stagewise' setting. We compare the bound and running time against the following approaches: \textbf{BDD+}, the Lagrangian Decomposition procedure proposed in \cite{Bunel2020-cl}, where the optimization is performed using optimal proximal step sizes; \textbf{LP}, the original linear programming relaxation using the PLANET relaxation on each ReLU neuron \cite{Zico_Kolter2017-va}; \textbf{Anderson}, the linear programming relaxation, where 1-cut is applied per neuron as in \cite{anderson2020strong}; \textbf{AS}, a recent Lagrangian technique which iteratively chooses cuts to employ from the Anderson relaxation \cite{De_Palma2021-sp}. 

\begin{wraptable}{r}{7cm}
\centering
\vspace{-4mm}
\caption{Timing data for the single-stage bounds.}\label{table:single-stage-time}
\begin{tabular}{lrr}
\toprule
Alg\textbackslash{}Net & MNIST Deep & CIFAR SGD \\
\midrule
BDD+ & \(2.5 \pm 0.2\) & \(1.5 \pm 0.1\) \\
LP & \(4.8 \pm 0.5\) & \(6.2 \pm 0.6\) \\
\textbf{ZD-2D} & \(10.9 \pm 0.5\) & \(7.6 \pm 0.3\) \\
AS & \(16.9 \pm 1.2\) & \(10.2 \pm 0.8\) \\
\textbf{ZD-MIP} & \(23.1 \pm 12.4\) & \(11.0 \pm 3.1\) \\
Anderson & \(52.4 \pm 27.5\) & \(21.3 \pm 4.6\) \\
\bottomrule
\end{tabular}
\end{wraptable}
All baselines are run implemented using the code and hyperparameter choices from prior works, and network weights are taken from existing works where applicable. Architecture descriptions are contained in the supplementary. We follow \cite{Bunel2020-cl} to construct scalar-valued verification problems: for examples with label $i$, we compare the difference between the $i^{th}$ and $(i+1)^{th}$ logit of a network. We can extend this procedure to a multi-class task by applying the verification $n-1$ times for each incorrect logit.

\paragraph{Single-stage bound evaluation:} In the single-stage setting we make the following hyperparameter choices for our approach. We initialize the intermediate bounds using zonotopes informed by the boxes attained by the best of the $\texttt{DeepZ}$ bounds and the IBP bounds. We then decompose each zonotope into 2-dimensional partitions, using the `similarity' heuristic for fully-connected networks, and the `spatial' heuristic for convolutional networks. Then we perform 1000 iterations of the Adam optimizer during the dual ascent procedure. We report the value at the last iteration as \textbf{ZD-2D}. Then we merge the partitions of only the last layer of each network into 20-dimensional zonotopes, and evaluate the dual function under this new partitioning. We report these values as \textbf{ZD-MIP}. 

We present results of our approach on the MNIST-Deep network, which has 5196 neurons and 6 layers, evaluating over a domain of $\ell_\infty$ boxes with $\epsilon=0.1$. Specifically we present cumulative density function plots: a point $(x,y)$ on any curve in these plots indicates that that particular algorithm attains a bound tighter than $x$ for $(1-y)\%$ of the examples. Curves further to the right indicate tighter bounds in aggregate. Figure \ref{fig:single-stage} (left) contains the distribution of reported bounds over every correctly-classified example in the MNIST validation set. In Figure \ref{fig:single-stage} (right), we present a similar CDF plot for a CIFAR-10 network with 6244 ReLU neurons and 4 layers. For this CIFAR network, we set $\epsilon=5/255$, as in \cite{De_Palma2021-sp}. The average and standard deviation runtimes for these networks is provided in Table \ref{table:single-stage-time}.

\begin{figure}[!htb]
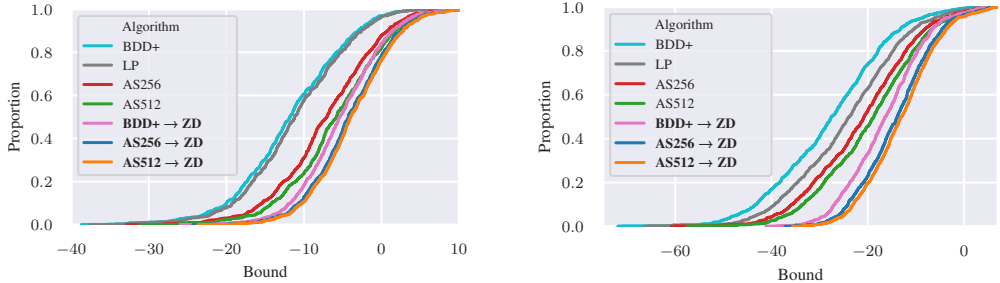

\vspace{-0.5ex}
\begin{subfigure}{0.5\textwidth}
  \centering
    \resizebox{0.9\textwidth}{!}{\input{figures/mnist_ffnet_all_cdf.pgf}}
\end{subfigure}
\begin{subfigure}{0.5\textwidth}
  \centering
    \resizebox{0.9\textwidth}{!}{\input{figures/mnist_deep_all_cdf.pgf}}
\end{subfigure}
\vspace{-1ex}
\caption{Stagewise bounds on the MNIST-FFNet (left) and MNIST-Deep (right) networks.}\label{fig:mnist-stagewise}
\end{figure}
\paragraph{Stagewise bound evaluation:} Now we turn our attention to the setting where enough computation time is allowed to perform the optimization procedure on every neuron in turn, which is able to generate tighter box bounds.  Here we notice an interesting observation: our method can be improved by leveraging tighter intermediate box bounds, but is agnostic as to where these bounds come from. Indeed, our approach can take as input any collection of tighter box bounds and generate tighter zonotopes via Proposition \ref{prop:box-deepz}. 
However, our current implementation of this tighter primal bound is not vectorized to be amenable, in the way that others are, to stagewise bounding procedures. This is purely an implementation detail, and we instead use existing primal bounding techniques.

As baselines here, we compare against the LP approach, BDD+, and the active set approach with 256 and 512 iterations, denoted AS256, AS512. We leverage box bounds provided by each of these dual approaches to inform our zonotope bounds from which we can run our method, denoted by a $\rightarrow \text{ZD}$. Note that BDD+ can never surpass bounds provided by the LP approach, whereas ours and the active set procedure can both improved even further with a longer runtime. Hence it is not sufficient to provide a better bound, but to also do so more efficiently. In this case, our approach performs 2000 iterations of Adam on 2-D zonotopes and increases the partition size of the final two layers of each network to sizes of 16 and 20 respectively.

\begin{wraptable}{r}{7.25cm}
\centering
\vspace{-4mm}
\caption{Timing data for stagewise bounds}
\centering
\begin{tabular}{lrr}
\toprule
Alg\textbackslash{}Net & MNIST Deep & MNIST Wide \\
\midrule
BDD+ & \(15.4 \pm 0.4\) & \(8.3 \pm 0.3\) \\
\textbf{BDD+ → ZD} & \(51.7 \pm 14.0\) & \(36.4 \pm 6.8\) \\
AS256 & \(138.9 \pm 0.9\) & \(45.4 \pm 0.6\) \\
\textbf{AS256 → ZD} & \(174.1 \pm 13.6\) & \(73.3 \pm 6.6\) \\
AS512 & \(289.3 \pm 1.7\) & \(91.3 \pm 0.9\) \\
\textbf{AS512 → ZD} & \(324.1 \pm 13.5\) & \(119.2 \pm 6.9\) \\
LP & \(2132.2 \pm 494.4\) & \(387.1 \pm 65.5\) \\
\bottomrule
\end{tabular}
\label{table:all-stage-mini}
\end{wraptable}

We present our results on the MNIST-FFNet and MNIST-Deep network which have 960 and 5196 neurons, and 5 and 6 layers respectively, again using an $\epsilon=0.1$. CDF plots of bounds for these two networks are provided in \ref{fig:mnist-stagewise}, with timing data in \ref{table:all-stage-mini}. We observe that the BDD+ bound completes very quickly, but attains a bound only slightly worse than LP. Our approach, when initialized with the BDD+ bounds runs more efficiently than AS256, providing bounds that are comparable or better than the slower AS512 approach. 

\paragraph{\%-Verified Robustness: } Finally we evaluate our approach comparing against all incorrect logits on MNIST trained networks, mirroring the setup of \cite{wang2021beta}. We report numbers from an unbranched version of ZD and BDD+ alongside several other verification techniques \cite{zhang2018efficient, muller2021precise, Wang2018-yb}. Compared to other efficient, branch-friendly procedures such as BDD+ and CROWN, our approach provides tighter bounds at a fair cost to efficiency. Our approach provides weaker bounds than PRIMA, but is significantly faster. $\beta$-CROWN, in particular, leverages a dual-based branching strategy which could be applied to our method. However due to its extremely efficient bounding procedure and the need to solve millions of subproblems for verification, it is unclear if our approach can surpass this method. 
\begin{table}[]
\caption{Verified Robustness on MNIST Networks. Methods denoted with an asterisk had numbers taken from \cite{wang2021beta}.} 
\label{table:pct-robust}
\begin{tabular}{l|c|l|l|l|l|l}
Model     & $\epsilon$ & \textbf{ZD}         & BDD+         & CROWN*      & PRIMA*        & $\beta$-CROWN* \\ \hline
MLP 5x100 & 0.026    & 28.5\% (24s) & 19.8\% (7s)  & 16.0\% (1s) & 51.0\% (159s) & 69.9\% (102s)      \\
MLP 8x100 & 0.026    & 21.8\% (19s) & 19.1\% (15s) & 18.2\% (1s) & 42.8\% (301s) & 62.0\% (103s)      \\
MLP 5x200 & 0.015    & 41.8\% (28s) & 33.2\% (8s)  & 29.2\% (2s) & 69.0\% (224s) & 77.4\% (86s)       \\
MLP 8x200 & 0.015    & 29.7\% (26s) & 27.4\% (17s) & 25.9\% (6s) & 62.4\% (395s) & 73.5\% (95s)       \\
ConvSmall & 0.12     & 52.6\% (38s) & 16.6\% (6s)  & 15.8\% (3s) & 59.8\% (42s)  & 72.7\% (7s)       
\end{tabular}
\end{table}

\section{Conclusion}
In this work, we demonstrated how primal and dual verification  techniques can be combined to yield efficient bounds that are tighter than either of them alone. Primal methods capture complex neuron dependencies, but do not optimize for a verification instance; while dual methods are highly scalable, but rely on unnecessarily loose primal bounds. This work is a first step in combining these two approaches. We believe that this combination of approaches is a necessary direction for providing scalable and tight robustness guarantees for safety-critical applications of machine learning.

\paragraph{Acknowledgements:} This work is supported in part by NSF grants CNS-2002664, DMS-2134012, CCF-2019844 as a part of NSF Institute for Foundations of Machine Learning (IFML), CNS-2112471 as a part of NSF AI Institute for Future Edge Networks and Distributed Intelligence (AI-EDGE), CCF-1763702,
AF-1901292, CNS-2148141, Tripods CCF-1934932, and research gifts by Western Digital, WNCG IAP, UT Austin Machine Learning Lab (MLL), Cisco and the Archie Straiton Endowed Faculty Fellowship.


\FloatBarrier
\newpage


\section*{Checklist}

The checklist follows the references.  Please
read the checklist guidelines carefully for information on how to answer these
questions.  For each question, change the default \answerTODO{} to \answerYes{},
\answerNo{}, or \answerNA{}.  You are strongly encouraged to include a {\bf
justification to your answer}, either by referencing the appropriate section of
your paper or providing a brief inline description.  For example:
\begin{itemize}
  \item Did you include the license to the code and datasets? \answerYes{See Section~\ref{gen_inst}.}
  \item Did you include the license to the code and datasets? \answerNo{The code and the data are proprietary.}
  \item Did you include the license to the code and datasets? \answerNA{}
\end{itemize}
Please do not modify the questions and only use the provided macros for your
answers.  Note that the Checklist section does not count towards the page
limit.  In your paper, please delete this instructions block and only keep the
Checklist section heading above along with the questions/answers below.

\begin{enumerate}

\item For all authors...
\begin{enumerate}
  \item Do the main claims made in the abstract and introduction accurately reflect the paper's contributions and scope?
    \answerYes{}
  \item Did you describe the limitations of your work?
    \answerYes{See Appendix}
  \item Did you discuss any potential negative societal impacts of your work?
    \answerYes{See Appendix}
  \item Have you read the ethics review guidelines and ensured that your paper conforms to them?
    \answerYes{}
\end{enumerate}

\item If you are including theoretical results...
\begin{enumerate}
  \item Did you state the full set of assumptions of all theoretical results?
    \answerYes{}
        \item Did you include complete proofs of all theoretical results?
    \answerYes{See Appendices}
\end{enumerate}

\item If you ran experiments...
\begin{enumerate}
  \item Did you include the code, data, and instructions needed to reproduce the main experimental results (either in the supplemental material or as a URL)?
    \answerYes{Zipped in appendix}
  \item Did you specify all the training details (e.g., data splits, hyperparameters, how they were chosen)?
    \answerYes{Described in Appendix}
        \item Did you report error bars (e.g., with respect to the random seed after running experiments multiple times)?
    \answerNo{Standard deviations for times, but n/a for CDF plots}
        \item Did you include the total amount of compute and the type of resources used (e.g., type of GPUs, internal cluster, or cloud provider)?
    \answerYes{Compute environment delineated in appendix}
\end{enumerate}

\item If you are using existing assets (e.g., code, data, models) or curating/releasing new assets...
\begin{enumerate}
  \item If your work uses existing assets, did you cite the creators?
    \answerYes{Models and choices of $\epsilon$ given proper credit}
  \item Did you mention the license of the assets?
    \answerNo{}
  \item Did you include any new assets either in the supplemental material or as a URL?
    \answerNo{}
  \item Did you discuss whether and how consent was obtained from people whose data you're using/curating?
    \answerNA{}
  \item Did you discuss whether the data you are using/curating contains personally identifiable information or offensive content?
    \answerNo{MNIST/CIFAR only}
\end{enumerate}

\item If you used crowdsourcing or conducted research with human subjects...
\begin{enumerate}
  \item Did you include the full text of instructions given to participants and screenshots, if applicable?
    \answerNA{}
  \item Did you describe any potential participant risks, with links to Institutional Review Board (IRB) approvals, if applicable?
    \answerNA{}
  \item Did you include the estimated hourly wage paid to participants and the total amount spent on participant compensation?
    \answerNA{}
\end{enumerate}

\end{enumerate}

\newpage
\appendix

\section{Broader Impacts}
Our work focuses on proving robustness and security for machine learning applications. While any new technology may be used maliciously, we believe the technical advances proposed in this work are as benign as possible and will only serve to offer certification of existing technologies. 

\section{Limitations and Future Work}
While ZonoDual has some benefits over prior works, it is not without limitations. We enumerate these here, as well as potential ways to overcome these:
\begin{itemize}
    \item \textbf{Reliance on ReLU nonlinearities:} ZonoDual relies on the ability to solve ReLU programs over zonotopes using MIPs. Unfortunately, this approach cannot generalize to nonlinearites that are not piecewise linear, such as the sigmoid activation. While it might be possible to solve `sigmoid programs' over 2-dimensional zonotopes, the formulation becomes significantly more complicated. 
    \item \textbf{MIPs are not conducive to GPU acceleration:} A crucial component of our procedure is to evaluate the dual using a tighter relaxation during the final phase of ZonoDual. Prior works that leverage box bounds do not include such a tightening step, but we note that MIPs can not benefit from GPU computations. In practice, small MIPs can solve very quickly, but for extremely large nets, such as SOTA image models, it is unlikely that this approach can scale as nicely as other techniques. This presents a need for fast and tight lower bounds for the reduced verification problem when the network is known to only have one hidden layer. Unfortunately, such methods, such as SDP's cannot currently scale to the case where this would be helpful for us. 
    \item\textbf{Intermediate zonotopes may have very large order:} One drawback of using zonotopic intermediate bounds is that the number of generators can grow to be quite large for deeper layers in the network. This can slow down the dual computation procedure, even in the 2-dimensional case for which dual evaluation time is linear in the number of generators. There is room for improvement here in what is known as 'order reduction' techniques for zonotopes, which soundly compresses the number of generators of a zonotope. Naive approaches are very quick and can allow for a 2x compression factor with only marginal degradations to dual function values, but significantly tighter compressions require much more work. We have not included any of these accelerations in our presentation for the sake of simplicity.
    \item \textbf{Exact evaluations of the dual seem to be necessary for informative gradients:} One might hope that the dual need not be evaluated exactly during the iteration phase of ZonoDual. Such a procedure could allow for much faster and tighter dual ascent iterations when only informative gradients are required. Unfortunately the approaches we have tried in this domain (including Frank-Wolfe, L-BFGBS, and a modified simplex algorithm) do not provide gradients that are informative enough to yield good dual values for the final evaluation phase. It is possible there are techniques here that could work, but we have not found them. 
\end{itemize}

\section{Zonotope Pushforward Operators}
Here we describe the sound zonotope pushforward operators we use to generate intermediate bounds. Since zonotopes are closed under affine transformation, all that's required is to describe the pushforward operator for the ReLU operator. Specifically, given a zonotope $Z(c,E)$, we seek to generate a zonotope $Z(c', E')$ such that $\setfilter{\sigma(z)}{z\in Z(c,E)} \subseteq Z(c', E')$. First we describe the approach when no extra information is known. This was first presented in \cite{Mirman2018-nx}, but we adapt the analysis of \cite{Jordan2021-lc}. Next we consider the case when we also have the extra information in the form of a hyperbox $H(l,u):= \setfilter{x}{l\leq x \leq u}$, where the goal is to construct $Z(c',E')$ such that $\setfilter{\sigma(z)}{z\in Z(c,E)\cap H(l,u)}\subseteq Z(c',E').$ 

\paragraph{\texttt{DeepZ}:}
The primary strategy is to compress each coordinate of $Z(c,E)$ by a scale factor $\lambda_i \in [0,1]$, which will in general, incur some error along each coordinate, $b_i$. These errors are accumulated into a hyperbox, which will then be incorporated into $Z(c', E')$ in the form of a Minkowski sum. The goal is to minimize the amount of error that is accumulated. In other words, each coordinate can be considered independently, where we wish to solve the minmax procedure
\[\min_{\lambda_i, b_i} \left(\max_{z\in Z(c,E)} |\sigma(z_i) - \lambda_iz_i - b_i|\right)\]
where the choice of the compression factor $\lambda_i$ is chosen to minimize the error that needs to be added in at the end between $\sigma(z_i)$ and $\lambda_i z_i$ over all feasible points in the zonotope. Noticing that projecting $Z(c,E)$ only onto the $i^{th}$ coordinate necessarily yields an interval, so the $\max$ is performed over an interval, $[l_i, u_i]$. In the case that $l_i \geq 0$, the ReLU is always on, and $\lambda_i$ can be chosen to be 1, with $b_i = 0$. In the case that $u_i \leq 0$, then the ReLU is always off, and $\lambda_i$ can be chosen to be 0, again with $b_i=0$. In the case that $l_i < 0 < u_i$, it is not difficult to see that the optimum value is attained when $\lambda_i$ is chosen to be $\frac{u_i}{u_i-l_i}$, such that the optimal value is $\frac{-l_iu_i}{u_i-l_i}$. Hence we can accumulate $\lambda_i$ for each coordinate $i$ into a diagonal matrix $\Lambda$. The optimal offsets, $(b_1,\dots, b_d)$, can be divided by 2, and accumulated into a diagonal matrix $B$ whose columns may be appended to the compressed zonotope. Put concisely, 
\[Z(c',E'):=Z(\Lambda c, \Lambda E |B )\]
where 
\[\begin{split}
    \Lambda_{i,i} = \begin{cases}
        0 & \text{ if } u_i \leq 0\\
        1 &\text{ if } l_i > 0\\
        \frac{u_i}{u_i-l_i} &\text{ otherwise }
    \end{cases}
    &\qquad 
    B_{i,i} = \begin{cases}
        0 & \text{ if } u_i \cdot l_i \geq 0 \\
        \frac{-l_i\cdot u_i}{2(u_i-l_i)} & \text{ otherwise}
    \end{cases}
\end{split}
\]

\paragraph{Improvements using Hyperboxes:}
Now we restate our proposition that allows for an improvement upon this pushforward operator when we have an incomparable hyperbox bound. 
\hboxProp* 
\begin{proof}
Let $Z(c,E)$ be the zonotope $\Z$, and let $H(l^h,u^h)$ be the hyperbox $\mathcal{H}$. Similar to \texttt{DeepZ}, the strategy we employ to develop a sound pushforward operator is to scale each coordinate independently, and Minkwoski sum the errors, accumulated as a hyperbox. Again, this decomposes into a minmax procedure as 
\[\min_{\lambda_i, b_i} \left(\max_{z\in Z(c,E)\cap H(l^h,u^h)} |\sigma(z_i) - \lambda_i z_i - b_i|\right). \]
In a single dimension, both a zonotope and hyperbox are intervals, so their intersection is also an interval. Letting $(l_i^z, u_i^z)$ be the $i^{th}$ coordinates lower and upper bounds for $\Z$, then the intersection between the $i^{th}$ coordinate projection of $\mathcal{H}$ and $\Z$ is the interval $[l_i, u_i] := [\max(\{l_i^z, l_i^h\}, \min(\{u_i^z, u_i^h\})]$. From here the choice of $\Lambda$ and $B$ is exactly the same as in \texttt{DeepZ}, where the tightened coordinate-wise bounds are used. This is guaranteed to be as tighter as long as at least one coordinate exists for which the interval is reduced, because both the compression factor $\lambda_i$ and error $b_i$ are reduced by tightening the interval. By our assumption that $\Z \centernot\subseteq \mathcal{H}$, at least one coordinate interval must be tightened, so the inclusion $\Z' \subset \texttt{DeepZ}(\Z)$ is strict.
\end{proof}

\begin{figure}
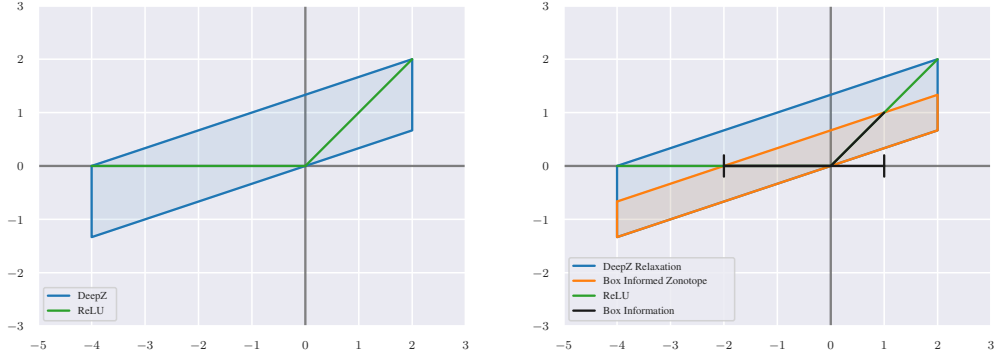

\centering
\begin{subfigure}{0.5\textwidth}
  \centering
    \resizebox{0.9\textwidth}{!}{\input{figures/deepZ_pgram.pgf}}
  \label{fig:deepz-pgram}
\end{subfigure}%
\begin{subfigure}{0.5\textwidth}
  \centering
    \resizebox{0.9\textwidth}{!}{\input{figures/deepZ_boxpgram.pgf}}
  \label{fig:deepz-boxpgram}
\end{subfigure}
\caption{Graphical representation of the coordinatewise bounding procedure of both \texttt{DeepZ} and our improved version when box bounds are available. On the left, the blue zonotope is the one introduced by \texttt{DeepZ}. On the right, we are also aware than our interval is bounded by the blue-interval, and the blue zonotope is the improved bound.}
\label{fig:deepz-grams}
\end{figure}

\begin{figure}
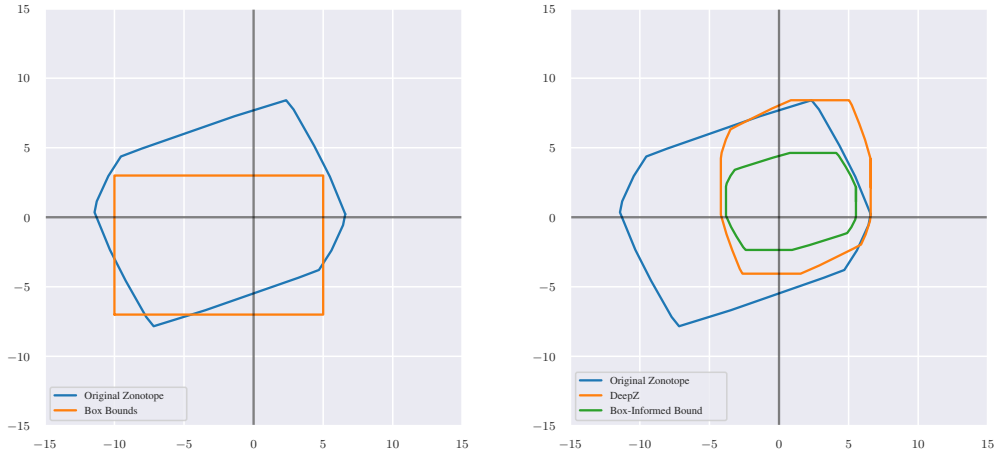

\centering
\begin{subfigure}{0.5\textwidth}
  \centering
    \resizebox{0.9\textwidth}{!}{\input{figures/deepZ_preReLU.pgf}}
  \label{fig:deepz-preReLU}
\end{subfigure}%
\begin{subfigure}{0.5\textwidth}
  \centering
    \resizebox{0.9\textwidth}{!}{\input{figures/deepZ_postReLU.pgf}}
  \label{fig:deepz-postReLU}
\end{subfigure}
\caption{An application of the improved zonotope pushforward operator on a 2-d example. On the left, the original zonotope is outlined in blue, and the extra hyperbox information is drawn in orange. On the right, the output of \texttt{DeepZ} is drawn in orange. The output of the improved pushforward operator, that has access to the box, is drawn in green.}
\label{fig:deepz-relus}
\end{figure}

\paragraph{Kolter-Wong Bounds}
Here we present a geometric proof of Proposition \ref{prop:zono-kw}. We will only briefly discuss the derivation of this bound in the context of this proof. For a more thorough discussion on the exact structure and derivation of the Kolter/Wong dual relaxation, we point to the original paper \cite{Zico_Kolter2017-va} and a self-contained derivation in the appendix of \cite{Bunel2020-cl}. 

\kwProp*
\begin{proof}
In \cite{Zico_Kolter2017-va}, the authors derive the original linear programming relaxation and dual procedure for verifying ReLU networks. This technique can be interpreted in the primal sense as maintaining a polyhedral primal bound for every intermediate layer. Assuming a polyhedral input range, such a hyperbox, this procedure works by incrementally lifting the dimensionality of the polytope. In other words, an H-polytope is maintained, where new dimensions/variables are introduced through equality constraints defined by each layer. For affine layers, this lifting procedure is exact. For ReLU layers, this lifting procedure introduces some error. This follows because the graph of the ReLU function, viewed as a 2-d set, is not convex. In this case the canonical triangle relaxation is applied, which is simply the convex hull of the graph of the ReLU function. This triangle relaxation, applied over an interval $[l,u]$ with $l < 0 < u$, has as its upper convex hull, a line with slope $\frac{u}{u-l}$ and intercept $\frac{-ul}{u-l}$. This lifting procedure is iterated over all layers, ultimately yielding a polytope of dimension $dim(\X) + \sum_{i=1}^L 2n_i$, where $n_i$ refers to the number of neurons of the $i^{th}$ layer. Optimization over this lifted polytope was framed as a linear program, where the objective was set to minimize the variable corresponding to the final output neuron.

In the original KW formulation, the dual of this linear program, a dual variable was introduced that geometrically corresponds to a lower-bounding linear functional on the triangle relaxation. This is similar to the observation made in $\alpha-\beta-CROWN$ \cite{wang2021beta}. In the original linear programming paper, an initial choice of $\alpha=\frac{u}{u-l}$ was chosen, which in this case may be interpreted as lower bounding the triangle relaxation by a line that is parallel to the upper convex hull, essentially providing an initial relaxation as a parallelogram with vertical sides. 

On the other hand, the zonotope propagation procedure may also be viewed as a polyhedral lifting procedure. Recall in the \texttt{DeepZ} formulation, the sound pushforward operator for the ReLU operator works by scaling the $i^{th}$ coordinate by the factor $\frac{u_i}{u_i-l_i}$ and incorporating the error in terms of a Minkowski sum. This may equivalently be viewed as lifting a $d$-dimensional zonotope, to a $(d+1)$-dimensional zonotope where the newly added dimension has range that is a function of only the $i^{th}$ coordinate. In particular, it is constrained to lie inside the vertical parallelogram defined on the vertical edges by $l, u$, with the slanted edge having slope $\frac{u_i}{u_i-l_i}$. The upper edge is exactly the convex upper hull of the graph of the ReLU function. In practice, because it is convenient and efficient to project zonotopes onto lower dimensional subspaces, when mapping a $d$-dimensional through a ReLU layer, instead of lifting to a $2d$-dimensional zonotope, the original dimensions are simply dropped to retain a $d$-dimensional zonotope. 

It is in this sense that the convex outer adversarial polytope described by Kolter and Wong, with the choice of $\alpha=\frac{u}{u-l}$ is exactly equivalent to the zonotope attained by \texttt{DeepZ} if no dimensions were dropped in the zonotope. Hence, each coordinate-wise upper and lower bound attained by each is exactly equivalent.
\end{proof}

\section{Proof of Hardness of ReLU Programming} 
\rphardness*
\begin{proof} 
Let $\Z = Z(c,E)$ be a zonotope in $\mathbb{R}^d$ with $m$ columns in the generator matrix. Then we can construct an equivalent scalar-valued ReLU network, $f(y)=w_2^\top\sigma(W_1y + b)$. We define $w_2, W_1, b$ as follows:
\[
\begin{split}
W_1 = \begin{bmatrix}  
E \\
-E 
\end{bmatrix}& \qquad 
\begin{split}
b = \begin{bmatrix} 
c \\
-c
\end{bmatrix} &\qquad
w_2 = \begin{bmatrix} c_1 + c_2\\-c1 \end{bmatrix}
\end{split}
\end{split}
\]
such that $f(y)=c_2^\top\sigma(c+Ey) + c_1^\top(\sigma(c+Ey) - \sigma(-c-Ey))T$, where the ReLU programming objective is recovered by $x=\sigma(x)-\sigma(-x)$. Then we can set the input domain to $[-1,1]^m$ to equate this verification problem to ReLU programming over $\Z$. The hardness result comes directly from Katz et al \cite{Katz2017-qz}. 
\end{proof}

\section{2-Dimensional Zonotopes}
Now we turn our attention to 2-dimensional zonotopes, ultimately, we will arrive at a proof of Theorem \ref{thm:2dzono-vertices} and Corollary \ref{cor:2d-zono-box}. 
Throughout, we consider a 2-dimensional zonotope $Z(c,E)$, where $E$ is assumed to have $m$ generators. 
Further we will assume that the generators are in general position, i.e. two generators are colinear. 
If this is the case they can be combined into a single generator without changing the set represented by the zonotope. 
The strategy of this proof will follow in several parts. 
First we start with a preliminary theorem that holds for zonotopes of all dimension, which describes a relationship between the vertices of the zonotope and the faces of the generating hypercube. 
Then we show how every two-dimensional zonotope has $2m$ vertices and therefore $2m$ edges. 
Next we prove that each edge corresponds to exactly one generator, and conversely, every generator corresponds to exactly 2 edges. Finally we describe the procedure we use to enumerate these vertices in a sorted fashion, and compute axis-crossings, origin-containment, and extend this to the intersection of a zonotope and rectangle. We will commonly make use of the notion of the \emph{upper and lower convex hulls} of a 2-dimensional shape. The upper hull of a convex set is the smallest concave function that is larger than the convex hull, and the lower convex hull of a convex set is the largest convex function that is smaller than the convex hull; it is helpful to think of this as the `upper half` and `lower half' of the convex hull. 
\begin{lemma}
Every vertex $v$ of a zonotope with $m$ generators has a corresponding face of the $m$-dimensional $\ell_\infty$ ball that maps to that vertex.
\end{lemma}
\begin{proof}
Consider any zonotope $\Z=Z(c,E)$, which can be equivalently viewed as the affine mapping $y\to c+Ey$ applied to every element in the set $\setfilter{y}{ y \in [-1, 1}^m\}$. 
Consider any vertex $v$ of $\Z$, which by definition must have some vector $a$ (in the polar cone of that vertex), such that $a^\top v < a^\top z$ for all $z\neq v$ in $\Z$. 
In other words, $v$ is the argmin of some linear program with feasible set $\Z$. 
Equivalently, the set of $y$ such that $c+Ey=v$ are the argmin of some linear program over the $\ell_\infty$ ball in $\mathbb{R}^m$. This implies that the set $\setfilter{y}{c+Ey=v}$ is a face of the $\ell_\infty$ ball. Note that face here includes 0-faces such as vertices of the $\ell_\infty$ ball, as well as higher dimensional faces. 
\end{proof}

\begin{lemma}\label{lemma:2m-edges}
A 2-dimensional zonotope with $m$ generators in general position has exactly $2m$ vertices and edges.
\end{lemma}
\begin{proof}
We proceed by induction. First note that a zonotope with a two generators is a parallelogram in $2d$ and has  4 vertices and edges. 
Now assume we have a zonotope $\Z_k$ with $k$ generators, and by the induction hypothesis, $2k$ vertices. 
Now assume without loss of generality that we add the $k+1^{th}$ generator which is the vector $g:=[0,1]$: we are always able to rotate and scale the zonotope such that this is the case, and invert this operation afterwards. If $V_k=\{v_1, \dots, v_{2k}\}$ is the set of vertices of $\Z_k$, then $\Z_k$ can be equivalently represented as $\operatorname{conv}(V_k)$. Because adding a generator to a zonotope is the minkowski sum of a segment, $\Z_{k+1}$ can equivalently be written as $\operatorname{conv}(\tilde{V_{k+1}})$, where $\tilde{V_{k+1}}$ is the set of points $\{v_1+g, v_1-g, \dots v_{2k}+g, v_{2k}-g\}$. We now show that only $2k+2$ of these $4k$ points are vertices. To see this, consider just the vertices of the convex upper hull of $\Z_k$. Because zonotopes are centrally symmetric, there are exactly $k+1$ of these points. By choosing $g$ to be $[0,1]$, we see that for every $v_i$ in the convex upper hull, $v_i+g$ is a vertex of $\Z_{k+1}$. We also see that unless $v_i$ is also a point of the convex lower hull of $\Z_k$, $v_i-g$ is not a vertex of $\Z_{k+1}$, because by definition there is an interval vertically beneath $v_i$ such that the entire interval, with $g$ subtracted, is also in $\Z_{k+1}$. Hence for $k-1$ points of the convex upper hull, we can eliminate $v_{i}-g$ from the candidate vertex set. We can repeat a similar procedure for the convex lower hull. Hence we are able to eliminate $2k-2$ points from the $4k$ candidate vertices, leaving exactly $2k+2$ vertices in $\Z_k$ as desired. Finally notice that every 2-d polytope with $2k+2$ vertices has exactly $2k+2$ edges.
\end{proof}

\definecolor{ggb_blue}{rgb}{0.08235294117647059,0.396078431372549,0.7529411764705882}
\definecolor{ggb_black}{rgb}{0.3803921568627451,0.3803921568627451,0.3803921568627451}
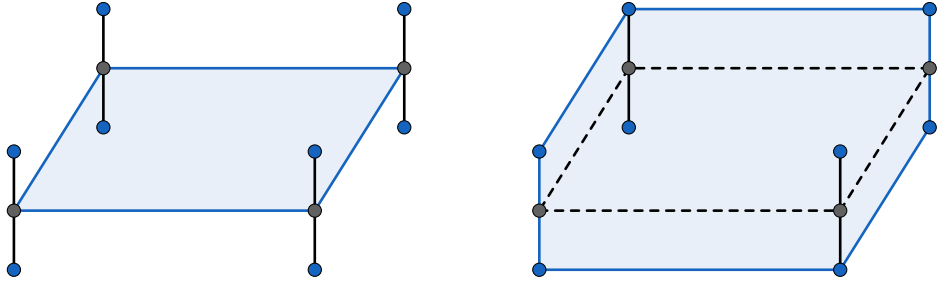
\begin{figure}
  \centering
  \begin{subfigure}{0.5\textwidth}
    \centering
    \begin{tikzpicture}[line cap=round,line join=round,x=1cm,y=1cm]
    \fill[line width=1pt,color=ggb_blue,fill=ggb_blue,fill opacity=0.10000000149011612] (0,0) -- (1.189462896759327,1.8948291271445854) -- (5.189462896759327,1.8948291271445854) -- (4,0) -- cycle;
    \draw [line width=1pt,color=ggb_blue] (0,0)-- (1.189462896759327,1.8948291271445854);
    \draw [line width=1pt,color=ggb_blue] (1.189462896759327,1.8948291271445854)-- (5.189462896759327,1.8948291271445854);
    \draw [line width=1pt,color=ggb_blue] (5.189462896759327,1.8948291271445854)-- (4,0);
    \draw [line width=1pt,color=ggb_blue] (4,0)-- (0,0);
    \draw [line width=1pt] (0,0)-- (0,0.7867211032548992);
    \draw [line width=1pt] (0,0)-- (0,-0.7867211032548992);
    \draw [line width=1pt] (1.189462896759327,1.8948291271445854)-- (1.189462896759327,2.6815502303994845);
    \draw [line width=1pt] (1.189462896759327,1.8948291271445854)-- (1.189462896759327,1.1081080238896863);
    \draw [line width=1pt] (5.189462896759327,1.8948291271445854)-- (5.189462896759327,2.6815502303994845);
    \draw [line width=1pt] (5.189462896759327,1.8948291271445854)-- (5.189462896759327,1.1081080238896868);
    \draw [line width=1pt] (4,0)-- (4,0.7867211032548993);
    \draw [line width=1pt] (4,0)-- (4,-0.7867211032548992);
    \begin{scriptsize}
    \draw [fill=ggb_black] (4,0) circle (2.5pt);
    \draw [fill=ggb_black] (1.189462896759327,1.8948291271445854) circle (2.5pt);
    \draw [fill=ggb_black] (5.189462896759327,1.8948291271445854) circle (2.5pt);
    \draw [fill=ggb_black] (0,0) circle (2.5pt);
    \draw [fill=ggb_blue] (0,0.7867211032548992) circle (2.5pt);
    \draw [fill=ggb_blue] (0,-0.7867211032548992) circle (2.5pt);
    \draw [fill=ggb_blue] (1.189462896759327,1.1081080238896863) circle (2.5pt);
    \draw [fill=ggb_blue] (1.189462896759327,2.6815502303994845) circle (2.5pt);
    \draw [fill=ggb_blue] (5.189462896759327,1.1081080238896868) circle (2.5pt);
    \draw [fill=ggb_blue] (5.189462896759327,2.6815502303994845) circle (2.5pt);
    \draw [fill=ggb_blue] (4,-0.7867211032548992) circle (2.5pt);
    \draw [fill=ggb_blue] (4,0.7867211032548993) circle (2.5pt);
    \end{scriptsize}
    \end{tikzpicture}
  \label{fig:2m-edges:before}
  \end{subfigure}%
  \begin{subfigure}{0.5\textwidth}
    \centering
    \begin{tikzpicture}[line cap=round,line join=round,x=1cm,y=1cm]
    \fill[line width=1pt,color=ggb_blue,fill=ggb_blue,fill opacity=0.10000000149011612] (0,0.7867211032548992) -- (1.189462896759327,2.6815502303994845) -- (5.189462896759327,2.6815502303994845) -- (5.189462896759327,1.1081080238896868) -- (4,-0.7867211032548992) -- (0,-0.7867211032548992) -- cycle;
    \draw [line width=1pt] (1.189462896759327,1.8948291271445854)-- (1.189462896759327,2.6815502303994845);
    \draw [line width=1pt] (1.189462896759327,1.8948291271445854)-- (1.189462896759327,1.1081080238896863);
    \draw [line width=1pt] (4,0)-- (4,0.7867211032548993);
    \draw [line width=1pt] (4,0)-- (4,-0.7867211032548992);
    \draw [line width=1pt,color=ggb_blue] (0,0.7867211032548992)-- (1.189462896759327,2.6815502303994845);
    \draw [line width=1pt,color=ggb_blue] (1.189462896759327,2.6815502303994845)-- (5.189462896759327,2.6815502303994845);
    \draw [line width=1pt,color=ggb_blue] (5.189462896759327,2.6815502303994845)-- (5.189462896759327,1.1081080238896868);
    \draw [line width=1pt,color=ggb_blue] (5.189462896759327,1.1081080238896868)-- (4,-0.7867211032548992);
    \draw [line width=1pt,color=ggb_blue] (4,-0.7867211032548992)-- (0,-0.7867211032548992);
    \draw [line width=1pt,color=ggb_blue] (0,-0.7867211032548992)-- (0,0.7867211032548992);
    \draw [line width=1pt,dash pattern=on 3pt off 3pt] (0,0)-- (4,0);
    \draw [line width=1pt,dash pattern=on 3pt off 3pt] (4,0)-- (5.189462896759327,1.8948291271445854);
    \draw [line width=1pt,dash pattern=on 3pt off 3pt] (5.189462896759327,1.8948291271445854)-- (1.189462896759327,1.8948291271445854);
    \draw [line width=1pt,dash pattern=on 3pt off 3pt] (1.189462896759327,1.8948291271445854)-- (0,0);
    \begin{scriptsize}
    \draw [fill=ggb_black] (4,0) circle (2.5pt);
    \draw [fill=ggb_black] (1.189462896759327,1.8948291271445854) circle (2.5pt);
    \draw [fill=ggb_black] (5.189462896759327,1.8948291271445854) circle (2.5pt);
    \draw [fill=ggb_black] (0,0) circle (2.5pt);
    \draw [fill=ggb_blue] (0,0.7867211032548992) circle (2.5pt);
    \draw [fill=ggb_blue] (0,-0.7867211032548992) circle (2.5pt);
    \draw [fill=ggb_blue] (1.189462896759327,1.1081080238896863) circle (2.5pt);
    \draw [fill=ggb_blue] (1.189462896759327,2.6815502303994845) circle (2.5pt);
    \draw [fill=ggb_blue] (5.189462896759327,1.1081080238896868) circle (2.5pt);
    \draw [fill=ggb_blue] (5.189462896759327,2.6815502303994845) circle (2.5pt);
    \draw [fill=ggb_blue] (4,-0.7867211032548992) circle (2.5pt);
    \draw [fill=ggb_blue] (4,0.7867211032548993) circle (2.5pt);
    \end{scriptsize}
    \end{tikzpicture}
  \label{fig:2m-edges:after}
\end{subfigure}
\caption{Pictorial aid for Lemma \ref{lemma:2m-edges}. We consider the following 2-dimensional shapes. The parallelogram on the left is the base case -- a zonotope with 2 generators, denoted by the convex hull of the gray vertices only. By incorporating a new vertical generator, the new candidate vertices are colored blue. Taking the convex hull of the blue vertices, we arrive at the 2-dimensional irregular hexagon on the right. Here we see that only the vertices on the left and right endpoints of the upper convex hull yield 2 vertices, whereas the other original vertices only generate 1 new vertex.}
\label{fig:2m-edges}
\end{figure}

\begin{lemma} \label{lemma:zono-generators}
If $\Z$ is a 2-dimensional zonotope in general position, then every edge corresponds to exactly one generator. 
Conversely, every generator corresponds to exactly two edges. 
\end{lemma}
\begin{proof}
Pick any edge of $\Z$ and without loss of generality, assume that zonotope is oriented such that the left endpoint of this edge, $v_l$, is the left-most point of the zonotope, neither edge connected to that endpoint is vertical, and the right endpoint of this edge, $v_r$ is above $v_l$. Note that this can always be done and then undone later. Next assume that all generators are oriented such that the $x$-coordinate of each generator is nonnegative. This follows because any generator can be negated without changing the zonotope. From the previous lemma, it can be shown that for zonotope $\Z=Z(c,E)$, the vector $y=-\vec{1}$ satisfies $c+Ey=v_l$. Then, by moving in $y$-space by swapping any coordinate of $y$ from $-1$ to $+1$, we may end up at a new vertex. We can now sort the generators by slope, in terms of $\frac{g(2)}{g(1)}$ , where $g(i)$ refers to the $i$\textsuperscript{th} coordinate of generator $g$. Suppose $g^{(j)}$ is the generator with the largest slope. Then by interpolating from $y$ to $y+2e_j$ we traverse along an edge of the zonotope. Indeed, there exists no other direction in $y$-space in which we could move that corresponds to a larger slope-increase in $z$-space. This is essentially a gift-wrapping procedure. Thus, the edge spanned by $v_l, v_r$ corresponds exactly to $g^{(j)}$. To show that every generator corresponds to at least two edges, this follows easily from the fact that zonotopes are centrally symmetric. To show that every generator corresponds to exactly two edges, we can apply Lemma \ref{lemma:2m-edges} to show by the pigeonhole principle, no generator can correspond to more than two edges. 
\end{proof}

Now with these lemmas in hand, we are prepared to solve the main 2-d vertex enumeration theorem. 
\zonoTwoD* 
\begin{proof}\label{thm:zono-enumeration}
First we describe how to enumerate all vertices of $Z(c,E)$ in $O(m\log m)$ time. This procedure is a simple extension of Lemma \ref{lemma:zono-generators}. Recall, in that lemma, we were able to find the left-most vertex by solving a linear program minimizing the $e_1^\top z$ over $Z(c,E)$. Then we were able to find the next vertex, starting from the left and going clockwise, by finding the generator with the largest slope, once generators had been appropriately sign-normalized. We can simply continue this procedure, and from the right-endpoint of the edge enumerated in Lemma \ref{lemma:zono-generators}, the next clockwise vertex is attained by traversing the generator with the next highest slope. In this way, we can sort the entire set of $m$ generators by their slope in decreasing order and iteratively enumerate the vertices, where the $(i+1)^{th}$ vertex $v_{i+1}$ is simply $v_{i} + 2g_i$, where $v_{i}$ is the $i^{th}$ vertex and $g_i$ is the $i^{th}$ generator, in the sorted order. This process generates the $m+1$ vertices of the entire convex upper hull of $Z(c, E)$ in left-right order. To attain the vertices of the lower convex hull, we rely on the centrally symmetric property of zonotopes: negate each vertex, add $2\cdot c$, and reverse the order of this list to generate the $m+1$ vertices of the lower convex hull in right-left order. By concatenating these two lists, and removing duplicates at the endpoints of each convex hull, we are able to generate all $2m$ vertices of the zonotope in clockwise order. The runtime of this algorithm is the time to compute the first vertex and then the slope of each generator, each of which is $O(m)$ time. Then the generator slopes must be sorted in $O(m\log m)$ time. Iterating through this list and adding each generator to the current vertex requires $O(m)$ constant time steps. Computing the lower convex hull only requires $O(m)$ time as well, so the whole procedure requires $O(m\log m)$ time, dominated by the cost of sorting the generators by their slopes.

Now we demonstrate how it takes an additional $O(m)$ time to compute the points on the zonotope where either $x=0$ or $y=0$. We consider only the $x=0$ case, without loss of generality. Since we are given the vertices in clockwise order, it is trivial to walk through the vertex list and report any pair of vertices for which $v_i(1) \cdot v_{i+1}(1) < 0$. If this is the case, then the edge between $v_i$ and $v_{i+1}$ crosses the $y$-axis. Since we have both endpoints on this edge, it requires constant time to compute the point along that line segment for which $x=0$. A similar procedure can be applied for the $x=0$ case. Finally, notice that the zonotope only contains the origin if it crosses both the $x$-axis and $y$-axis in two places. 
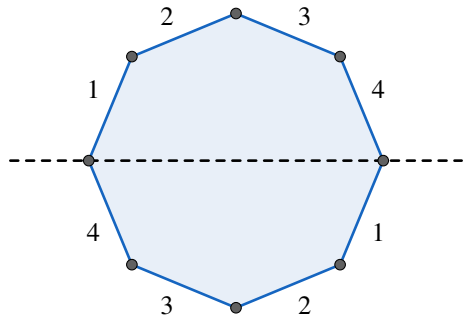
\begin{figure}[h]
    \centering
    \begin{tikzpicture}[line cap=round,line join=round,x=0.5cm,y=0.5cm]
    \fill[line width=1pt,color=ggb_blue,fill=ggb_blue,fill opacity=0.10000000149011612] (0,3.9148507336746783) -- (-2.768217501114496,2.768217501114496) -- (-3.9148507336746783,0) -- (-2.768217501114496,-2.768217501114496) -- (0,-3.9148507336746783) -- (2.768217501114496,-2.768217501114496) -- (3.9148507336746783,0) -- (2.768217501114496,2.768217501114496) -- cycle;
    \draw [line width=1pt,color=ggb_blue] (0,3.9148507336746783)-- node [black, midway, auto, swap] {2} (-2.768217501114496,2.768217501114496);
    \draw [line width=1pt,color=ggb_blue] (-2.768217501114496,2.768217501114496)-- node [black, midway, auto, swap] {1} (-3.9148507336746783,0);
    \draw [line width=1pt,color=ggb_blue] (-3.9148507336746783,0)-- node [black, midway, auto, swap] {4} (-2.768217501114496,-2.768217501114496);
    \draw [line width=1pt,color=ggb_blue] (-2.768217501114496,-2.768217501114496)-- node [black, midway, auto, swap] {3} (0,-3.9148507336746783);
    \draw [line width=1pt,color=ggb_blue] (0,-3.9148507336746783)-- node [black, midway, auto, swap] {2} (2.768217501114496,-2.768217501114496);
    \draw [line width=1pt,color=ggb_blue] (2.768217501114496,-2.768217501114496)-- node [black, midway, auto, swap] {1} (3.9148507336746783,0);
    \draw [line width=1pt,color=ggb_blue] (3.9148507336746783,0)-- node [black, midway, auto, swap] {4} (2.768217501114496,2.768217501114496);
    \draw [line width=1pt,color=ggb_blue] (2.768217501114496,2.768217501114496)-- node [black, midway, auto, swap] {3} (0,3.9148507336746783);
    \draw [line width=1pt,dash pattern=on 3pt off 3pt,domain=-6:6] plot(\x,{(-0-0*\x)/7.829701467349357});
    \begin{scriptsize}
    \draw [fill=ggb_black] (-3.9148507336746783,0) circle (2pt);
    \draw [fill=ggb_black] (3.9148507336746783,0) circle (2pt);
    \draw [fill=ggb_black] (0,-3.9148507336746783) circle (2pt);
    \draw [fill=ggb_black] (0,3.9148507336746783) circle (2pt);
    \draw [fill=ggb_black] (2.768217501114496,2.768217501114496) circle (2pt);
    \draw [fill=ggb_black] (-2.768217501114496,-2.768217501114496) circle (2pt);
    \draw [fill=ggb_black] (-2.768217501114496,2.768217501114496) circle (2pt);
    \draw [fill=ggb_black] (2.768217501114496,-2.768217501114496) circle (2pt);
    \end{scriptsize}
    \end{tikzpicture}
    \caption{Pictorial aid for Theorem \ref{thm:2dzono-vertices}. By only considering the edges on the upper convex hull (above the dotted line), we sort the edges according to slope and step along them until we reach the end of the upper convex hull. The convex lower hull can be computed easily by central symmetry of zonotopes.}
    \label{fig:zono-enumeration}
\end{figure}
\end{proof}

From this procedure, it is now trivial to consider the intersection of a 2-d zonotope and a rectangle. 
\zonoTwoDBox*
\begin{proof}

First we provide a bound on the cardinality of the set of vertices of the intersection of a 2-d zonotope and a rectangle. The set of candidate vertices is exactly: the vertices of the zonotope, the vertices of the rectangle, and any intersections of edges of the rectangle and zonotope. Here it is easiest to refer the pictorial aid in \cref{fig:zono-box-intersect}. For any rectangle vertex contained in the zonotope, there is necessarily a zonotope vertex not contained in their intersection. For any rectangle vertex not contained in the zonotope, it is possible that up to two new vertices are introduced in the intersection. This means that each vertex of the rectangle can add a net $+1$ to the number of vertices of their intersection. In general, the number of vertices of the intersection of two 2-d polygons is bounded by the sum of the number of vertices of each. Then the candidate optima for a ReLU program over this set is exactly this set of vertices, with cardinality at most $2m+4$, and any extra axis crossings, and the origin. This brings the set of candidate optima to no more than $2m+9.$

Next, we describe how to compute this set of candidate optima. We assume that we have obtained the $2m$ vertices of the zonotope in $O(m\log m)$ time. We now check containment of each of the candidate vertices. It requires constant time to compute point-containment within a rectangle, so we can eliminate any zonotope vertices not contained in the box in $O(2m)$ time. It is more complicated to compute point-containment within a zonotope. In general this requires a linear program, but since we already have the vertex list oriented clockwise, we are able to leverage this to compute the set of rectangle vertices contained in the zonotope as well as any edge-edge crossings in $O(m)$ time. The idea is similar to the idea used to compute axis crossings in Theorem \ref{thm:2dzono-vertices}.

The general procedure is to fix an axis and value and shoot a line through that value. For example, we consider the $x=a$ line and compute the intersection of the set $\setfilter{(x,y)}{x=a}$ with $Z(c,E)$. We can do this by finding the $y$-coordinates of the edges of $Z(c,E)$ that cross this vertical line. This amounts to a single scan through the oriented vertex list of the zonotope, and yields an interval $[l_z, u_z]$ or $\texttt{None}$. We perform this line-shooting procedure at most 6 times. Once for the left-bound, right-bound, top-bound, and bottom-bound of the rectangle, and if the rectangle itself crosses the $x$ or $y$ axis, we line-shoot the $x=0$ line or $y=0$ lines accordingly. For each line-shoot, we are able to compare the interval defined by the box $[l_b, u_b]$ and the line-shoot interval $[l_z, u_z]$. The intersection of this interval can be used to determine which box vertices and edge-edge intersections are contained within the intersection of the zonotope and rectangle. Overall this requires a constant number of calls to a subroutine running in $O(m)$ time, so the overall runtime of this procedure is still dominated by the sort time from the zonotope vertex enumeration, i.e. $O(m\log m).$

\begin{figure}[h]
    \centering
    \begin{tikzpicture}[line cap=round,line join=round,x=0.5cm,y=0.5cm]
    \fill[line width=2pt,color=ggb_blue,fill=ggb_blue,fill opacity=0.10000000149011612] (-1.530733729460359,3.695518130045147) -- (1.530733729460359,3.695518130045147) -- (3.695518130045147,1.530733729460359) -- (3.695518130045147,-1.530733729460359) -- (1.530733729460359,-3.695518130045147) -- (-1.530733729460359,-3.695518130045147) -- (-3.695518130045147,-1.530733729460359) -- (-3.695518130045147,1.530733729460359) -- cycle;
    \fill[line width=1pt,color=ggb_blue,fill=ggb_blue,fill opacity=0.10000000149011612] (-5,3) -- (3,3) -- (3,-1) -- (-5,-1) -- cycle;
    \draw [line width=1pt,color=ggb_blue] (-1.530733729460359,3.695518130045147)-- (1.530733729460359,3.695518130045147);
    \draw [line width=1pt,color=ggb_blue] (1.530733729460359,3.695518130045147)-- (3.695518130045147,1.530733729460359);
    \draw [line width=1pt,color=ggb_blue] (3.695518130045147,1.530733729460359)-- (3.695518130045147,-1.530733729460359);
    \draw [line width=1pt,color=ggb_blue] (3.695518130045147,-1.530733729460359)-- (1.530733729460359,-3.695518130045147);
    \draw [line width=1pt,color=ggb_blue] (1.530733729460359,-3.695518130045147)-- (-1.530733729460359,-3.695518130045147);
    \draw [line width=1pt,color=ggb_blue] (-1.530733729460359,-3.695518130045147)-- (-3.695518130045147,-1.530733729460359);
    \draw [line width=1pt,color=ggb_blue] (-3.695518130045147,-1.530733729460359)-- (-3.695518130045147,1.530733729460359);
    \draw [line width=1pt,color=ggb_blue] (-3.695518130045147,1.530733729460359)-- (-1.530733729460359,3.695518130045147);
    \draw [line width=1pt,color=ggb_blue] (-5,3)-- (3,3);
    \draw [line width=1pt,color=ggb_blue] (3,3)-- (3,-1);
    \draw [line width=1pt,color=ggb_blue] (3,-1)-- (-5,-1);
    \draw [line width=1pt,color=ggb_blue] (-5,-1)-- (-5,3);
    \begin{scriptsize}
    \draw [fill=ggb_black] (3.695518130045147,1.530733729460359) circle (2pt);
    \draw [fill=ggb_black] (1.530733729460359,3.695518130045147) circle (2pt);
    \draw [fill=ggb_black] (-1.530733729460359,3.695518130045147) circle (2pt);
    \draw [fill=ggb_black] (-3.695518130045147,1.530733729460359) circle (2pt);
    \draw [fill=ggb_black] (-3.695518130045147,-1.530733729460359) circle (2pt);
    \draw [fill=ggb_black] (-1.530733729460359,-3.695518130045147) circle (2pt);
    \draw [fill=ggb_black] (1.530733729460359,-3.695518130045147) circle (2pt);
    \draw [fill=ggb_black] (3.695518130045147,-1.530733729460359) circle (2pt);
    \draw [fill=ggb_black] (3,3) circle (2pt);
    \draw [fill=ggb_black] (-5,-1) circle (2pt);
    \draw [fill=ggb_black] (-5,3) circle (2pt);
    \draw [fill=ggb_black] (3,-1) circle (2pt);
    \draw [fill=ggb_blue] (3,2.226251859505505) circle (2pt);
    \draw [fill=ggb_blue] (2.226251859505505,3) circle (2pt);
    \draw [fill=ggb_blue] (-2.226251859505505,3) circle (2pt);
    \draw [fill=ggb_blue] (-3.695518130045147,-1) circle (2pt);
    \end{scriptsize}
    \end{tikzpicture}
    \caption{Pictorial aid for the proof of Lemma \ref{cor:2d-zono-box}. A zonotope (octagon) is intersected with a rectangle. Some rectangle vertices are not contained within the zonotope and can be eliminated, and vice versa. The rectangle vertex on the upper right introduces a net $+1$ vertices to the intersection.}
    \label{fig:zono-box-intersect}
\end{figure}
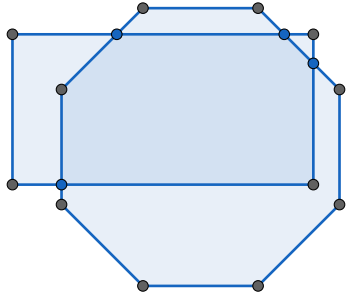
\end{proof}

\subsection{Partitioning Heuristics}
Here we describe the heuristics used to fragment a higher dimensional zonotope into two-dimnesional pieces. We certainly could do this randomly, but we have found some minor performance gains can be achieved by cleverly choosing this partitioning. The key idea for each of these is to choose pairs of dimensions yielding zonotopes that would suffer greatly from being relaxed into a rectangle. Recall that for a two-dimensional zonotope, a rectangle is attained when each generator is colinear with an elementary basis vector. Our two partitioning styles are as follows: 
\begin{itemize}
    \item \textbf{Similarity heuristic:} Here we assign each pair of dimensions a scalar-valued score that roughly resembles how far each generator column is from an elementary basis vector. These pairwise scores are encoded in a $d\times d$ matrix. Each row of this matrix is sorted in descending order, and the partitions are formed by looping over the rows of the matrix and choosing the first available index, that has not already been chosen by a prior pair.  The scoring function here, for dimensions $i,j$ is the dot product of the absolute values of the \emph{rows} of the generator matrix. Hence the scoring matrix can be computed by the matrix-matrix multiplication $|E|\cdot |E|^\top$. For zonotopes of very large dimension, this method requires the multiplication of two $(d\times m)$ matrices, and then the sort-and-choose method takes $O(d^2)$ time in the worst case. However since we only have to do this one time, it is often worth it to apply this heuristic on networks with intermediate networks with moderate dimension, or those without spatial information imbued by convolutional layers.
    \item \textbf{Spatial or depthwise heuristic:} For convolutional networks, the structure of convolutional operators can be applied to choose partitionings. In the spatial partitioning heuristic, we assign each coordinate its pair by choosing a coordinate adjacent to it in the feature map. In the depthwise partitioning heuristic, we assign each coordinate its pair by choosing a coordinate in the same location on the feature map but in a different channel. The intuition here is that random partitioning here would often choose coordinates that have disjoint receptive fields and would yield zonotopes that are rectangles. These heuristics much more quickly than the similarity heuristic and are amenable to larger convolutional nets. 
\end{itemize}

\section{Experimental Details}

\subsection{Computation Environment} 

Computations for this paper were performed using the idle resource queue of computing cluster. 
The cluster uses Intel Xeon Gold 6230R and 6230R CPUs and mixture of NVIDIA 2080 Ti, Quadro RTX 6000, and Telsa A40 GPUs.
Due to our usage of the idle resource queue, we ran on a variety of different CPUs and GPUs depending on which resources were being underutilized on the cluster. 
Our job requirements were 5 CPU cores, 20 GB of RAM, and 1 GPU of any kind per job.
Despite our timing information not being representative of a single hardware configuration, we believe we can still make valid comparisons using it, as we ensured that we ran all methods for a particular image on the same node.

\subsection{Network Architectures}

\subsubsection{MNIST Networks:}
We consider three different architectures trained on the MNIST dataset. Each is randomly initialized, and trained adversarially for 25 epochs using the Adam optimizer and standard hyperparameters to update network weights \cite{kingma2014adam}. The adversarial training was perfromed using  10 iterations of PGD with an $\ell_\infty$ adversarial radius of $\epsilon=0.1$ \cite{madry2017towards}. The architectures are as follows, where $\texttt{Conv2D}(in, out, k, s, p)$ refers to a convolutional layer with $in$ input channels, $out$ output channels, a square kernel of size $k$, a stride of $s$, and padding $p$. The Wide and Deep network architecture choices were taken from \cite{lu2019neural}, though the FFNet was our choice for a fully connected network. We trained these networks ourselves. 

\begin{itemize}
    \item \texttt{MNIST FFNet:} (960 ReLU neurons)
    \begin{enumerate}
        \item \(\texttt{Linear}(784, 512) \to \texttt{ReLU}\)
        \item \(\texttt{Linear}(512, 256) \to\texttt{ReLU}\)
        \item \(\texttt{Linear}(256, 128)\to \texttt{ReLU}\)
        \item \(\texttt{Linear}(128, 64)\to \texttt{ReLU}\)
        \item \(\texttt{Linear}(64, 10)\)
    \end{enumerate}
    \item \texttt{MNIST Wide:} (4804 ReLU neurons)
    \begin{enumerate}
        \item \(\texttt{Conv2D}(1, 16, 4, 2, 1) \to \texttt{ReLU}\)
        \item \(\texttt{Conv2D}(16, 32, 4, 2, 1) \to \texttt{ReLU}\to \texttt{Flatten}\)
        \item \(\texttt{Linear}(1568, 100) \to \texttt{ReLU}\)
        \item \(\texttt{Linear}(100, 10)\)
    \end{enumerate}
    \item \texttt{MNIST Deep:}  (5196 ReLU neurons)
    \begin{enumerate}
        \item \(\texttt{Conv2D}(1, 8, 4, 2, 1) \to \texttt{ReLU}\)
        \item \(\texttt{Conv2D}(8, 8, 3, 1, 1) \to \texttt{ReLU}\)
        \item \(\texttt{Conv2D}(8, 8, 3, 1, 1) \to \texttt{ReLU}\)
        \item \(\texttt{Conv2D}(8, 8, 4, 2, 1) \to \texttt{ReLU}\to \texttt{Flatten}\)
        \item \(\texttt{Linear}(392, 100) \to \texttt{ReLU}\)
        \item \(\texttt{Linear}(100, 10)\)
    \end{enumerate}
\end{itemize}
\subsubsection{CIFAR-10 Networks}
For the CIFAR-10 networks, we directly utilized the models and weights from prior work \cite{De_Palma2021-sp}\footnote{\href{https://github.com/oval-group/scaling-the-convex-barrier}{https://github.com/oval-group/scaling-the-convex-barrier}}. Specifically, we considered the models trained both using SGD and adversarially using PGD with the architecture: 
\begin{itemize}
\item \texttt{CIFAR WIDE:} (6244 ReLU neurons)
\begin{enumerate}
    \item \(\texttt{Conv2D}(3, 16, 4, 2, 1) \to \texttt{ReLU}\)
    \item \(\texttt{Conv2D}(16, 32, 4, 2, 1) \to \texttt{ReLU}\to \texttt{Flatten}\)
    \item \(\texttt{Linear}(2048, 100) \to \texttt{ReLU}\)
    \item \(\texttt{Linear}(100, 10)\)
\end{enumerate}
\end{itemize}
\subsection{Hyperparameter Choices}\label{app:hyperparameters}
Here we describe the hyperparameter choices for both our model and competing methods. Wherever possible, code from prior works was utilized, with minimal tuning applied. We found the codebase associated with the Active Set paper to be immensely helpful here \cite{De_Palma2021-sp}. Because we changed our partitioning hyperparameters slightly between networks, we describe each hyperparameter choice explicitly. Scripts to run each experiment are contained with the attached codebase. For all experiments, we employ Gurobi as  MIP-solver \cite{gurobi}. 
\paragraph{Single-stage setting:} 
\begin{itemize}
    \item \textbf{BDD+}: For the optimal proximal method, we apply the default parameters where 400 iterations were applied with 2 inner steps and $\eta$ ranging from 1 to 5. 
    \item \textbf{AS}: 1000 iterations of AS were increased to 1000 to match our approach, where all other hyperparameters were kept at their default values.
    \item \textbf{ZD (MNIST)}: For the MNIST networks, we initialized with the KW dual variables, used the 'similarity' partitioning heuristic for the 2-d zonotopes, and used 1000 iterations of Adam with default $\beta$ parameters. The learning rate was initialized at $\eta=0.01$, and decayed by a multiplicative $0.75$ factor every 100 iterations.
    \item \textbf{ZD (CIFAR)}: For the CIFAR networks, we initialized with the KW dual variables, used the 'spatial' partitioning heuristic for the 2-d zonotopes, and used 1000 iterations of Adam with default $\beta$ parameters. The learning rate was initialized at $\eta=0.0025$ and decayed by a multiplicative $0.5$ factor every 200 iterations. 
    \item \textbf{ZD-MIP}: For all single stage experiments, for every network, we ran the exact same setting as ZD, but merged partitions of the final layer only into zonotopes of dimension 20 and evaluated the dual function. The only exception here is in MNIST FFNET, where only the final layer was merged to zonotopes of dimension 16. 
\end{itemize}

\paragraph{Stagewise Setting:}
In this setting, we kept most hyperparameters the same as in the single-stage setting. The notable exception is a change in the number of iterations in the AS method. We found that 1000 iterations would frequently cause CUDA memory errors, and to the best of our understanding, these methods weren't run in a stagewise setting in \cite{De_Palma2021-sp}. Because these methods are allowed to run for a longer time, we increased the number of iterations dual ascent was performed for, as well as the partitioning at the end. Essentially, these were tuned slightly in order to reliably prove tighter bounds such that AS256-ZD was both faster and tighter than AS512. We summarize the iteration and partitioning changes here :
\begin{itemize}
    \item \textbf{MNIST FFNET}: We initialize as before, using the specified box bounds, and then run 2000 iterations of gradient ascent, with an initial learning rate of 0.0025, decaying by a factor of 0.75 every 400 epochs. Because this method is relatively smaller, we increase the final layer partition size to 32 and run 20 iterations of dual ascent. Finally, we expand all other 2-dimensional partitions to zonotopes of dimensin 16 and evaluate the dual.
    \item \textbf{MNIST WIDE/DEEP}: We initialize and run dual ascent over the 2-d zonotopes as in MNIST FFNet. We then increase partition sizes to 16 for the penultimate layer and 20 for the final layer and evaluate the dual function once. 
\end{itemize}

\paragraph{Ablation Study Details:} For the ablation study, we only evaluate on the MNIST FFNet. We initialize either with the \texttt{DeepZ} bounds, the \texttt{DeepZ} bounds augmented via IBP, or the zonotopes augmented by the stagewise BDD+ bounds. Then we partition the zonotopes using the `similarity' heuristic and run 1000 iterations of gradient ascent with an initial learning rate of 0.01, decaying by a factor of 0.75 every 100 iterations. 
We increase the partition size to 32 on the final layer only and evaluate the dual. 
We give results complete with error bars in \cref{table:ablation-full}.

\begin{table}[h]
\centering
\caption{Ablation study for each of phase of ZonoDual. We evaluate the bound and runtime after completion of the Initialization phase (init), Iteration phase (iter), and Evaluation (eval) phases, averaging over the first 1000 examples from MNIST.
We report three initialization techniques: DeepZ preactivation bounds, and DeepZ augmented with IBP or BDD+ hyperbox bounds.
The computed lower bounds are reported relative to the LP relaxation (a larger number means a tighter bound).}\label{table:ablation-full}
\begin{tabular}{lrrr}
\toprule
Init\textbackslash{}Phase & Init & Iter & Eval \\
\midrule
DeepZ & \(-3.9 \pm 1.1\) & \(2.9 \pm 1.0\) & \(7.7 \pm 2.6\) \\
Runtime (s) & \(0.0047 \pm 0.0004\) & \(5.6 \pm 0.3\) & \(7.0 \pm 1.5\) \\\midrule
DeepZ + IBP & \(-3.9 \pm 1.1\) & \(2.9 \pm 1.0\) & \(7.7 \pm 2.6\) \\
Runtime (s) & \(0.024 \pm 0.002\) & \(5.6 \pm 0.2\) & \(6.9 \pm 1.4\) \\\midrule
DeepZ + BDD+ & \(5.3 \pm 1.8\) & \(6.4 \pm 1.9\) & \(10.1 \pm 3.2\) \\
Runtime (s) & \(4.9 \pm 0.3\) & \(10.5 \pm 0.5\) & \(11.5 \pm 1.4\) \\
\bottomrule
\end{tabular}
\end{table}

\section{Additional experimental results}\label{appendix:exp-results}
Here we have included data from experiments that did not fit in the main paper. 
Unless otherwise stated, the setup and hyperparameters are chosen as in the main paper and in Appendix \ref{app:hyperparameters}.

\begin{figure}
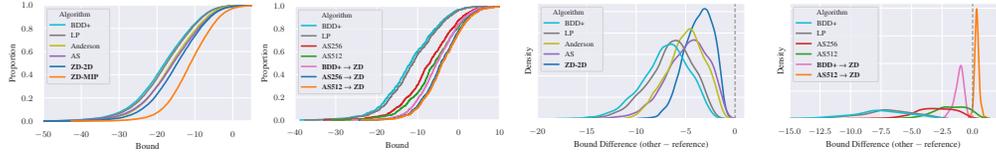

\centering
\begin{subfigure}{0.24\textwidth}
  \centering
    \resizebox{0.99\textwidth}{!}{\input{figures/mnist_ffnet_cdf.pgf}} 
  \caption{Single-stage CDF}\label{appfig:mnist-ffnet-cdf}
\end{subfigure}%
\begin{subfigure}{0.24\textwidth}
  \centering
 \resizebox{0.99\textwidth}{!}{\input{figures/mnist_ffnet_all_cdf.pgf}}
  \caption{All-stage CDF}\label{appfig:mnist-ffnet-all-cdf}
\end{subfigure}%
\begin{subfigure}{0.24\textwidth}
  \centering
  \resizebox{0.9\textwidth}{!}{\input{figures/mnist_ffnet_diff.pgf}} 
  \caption{Single-stage difference}\label{appfig:mnist-ffnet-diff}
\end{subfigure}%
\begin{subfigure}{0.24\textwidth}
  \centering
  \resizebox{0.9\textwidth}{!}{\input{figures/mnist_ffnet_all_diff.pgf}}
  \caption{All-stage difference}\label{appfig:mnist-ffnet-all-diff}
\end{subfigure}%
\caption{Full plots for the MNIST FFNet network (zoom in for detail). 
(\subref{appfig:mnist-ffnet-cdf}) The CDF plot for the single-stage setting. 
(\subref{appfig:mnist-ffnet-all-cdf}) The CDF plot for the multi-stage setting. 
(\subref{appfig:mnist-ffnet-diff}) The difference plot for the single-stage setting, where the reference method is ZD-MIP. (\subref{appfig:mnist-ffnet-all-diff}) The difference plot for the multi-stage setting, where the reference method is AS256$\to$ZD.
}
\label{appfig:mnist-ffnet}
\end{figure}

\begin{figure}
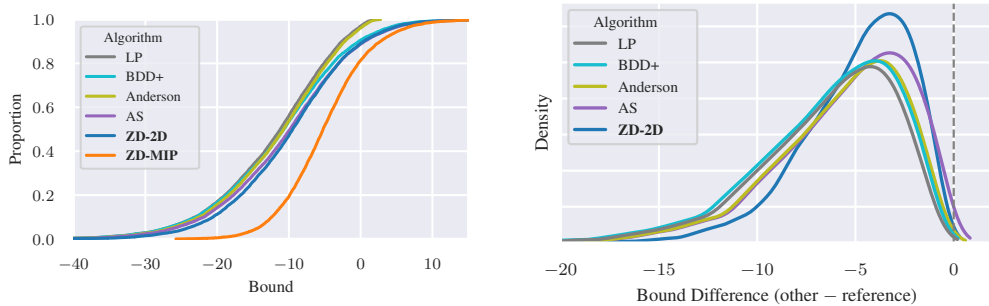

\centering
\begin{subfigure}{0.5\textwidth}
  \centering
    \resizebox{0.9\textwidth}{!}{\input{figures/cifar_sgd_cdf.pgf}}
  \caption{Single-stage CDF}\label{appfig:cifar-sgd-cdf}
\end{subfigure}%
\begin{subfigure}{0.5\textwidth}
  \centering
    \resizebox{0.9\textwidth}{!}{\input{figures/cifar_sgd_diff.pgf}}
  \caption{Single-stage difference}\label{appfig:cifar-sgd-diff}
\end{subfigure}
\caption{Plots for the CIFAR-10 network: (\subref{appfig:cifar-sgd-cdf}) CDF plots repeated from the main paper. (\subref{appfig:cifar-sgd-diff}) the difference plots where the reference method is the ZD-MIP method.}
\label{appfig:cifar-sgd}
\end{figure}

We present two styles of plots in \crefrange{appfig:mnist-ffnet}{appfig:cifar-sgd}.
The cumulative distribution plots present the quantity of examples considered that yield a bound less than a specified value. Curves further to the right are better here. The difference plots present the distribution of bound differences relative to a reference method, which is represented by the dotted vertical line. Curves further to the right provide better bounds, and portions of the curve greater than 0 indicate the proportion of examples that the method beats the reference. For example, in the single-stage setting, the curves represent the difference between the technique in the plot and the ZD-MIP method. In the multi-stage setting, we use AS256$\to$ZD as the reference.

We also report numerical timing results in \cref{table:single-stage-time-full,table:all-stage-time-full} and bound statistics in \cref{table:single-stage-bound-full,table:all-stage-bound-full}.

\begin{figure}
\centering
\begin{subfigure}{0.24\textwidth}
  \centering
    \resizebox{0.99\textwidth}{!}{\input{figures/mnist_wide_cdf.pgf}} 
  \caption{Single-stage CDF}\label{appfig:mnist-wide-cdf}
\end{subfigure}%
\begin{subfigure}{0.24\textwidth}
  \centering
 \resizebox{0.99\textwidth}{!}{\input{figures/mnist_wide_all_cdf.pgf}}
  \caption{All-stage CDF}\label{appfig:mnist-wide-all-cdf}
\end{subfigure}%
\begin{subfigure}{0.24\textwidth}
  \centering
  \resizebox{0.9\textwidth}{!}{\input{figures/mnist_wide_diff.pgf}} 
  \caption{Single-stage difference}\label{appfig:mnist-wide-diff}
\end{subfigure}%
\begin{subfigure}{0.24\textwidth}
  \centering
  \resizebox{0.9\textwidth}{!}{\input{figures/mnist_wide_all_diff.pgf}}
  \caption{All-stage difference}\label{appfig:mnist-wide-all-diff}
\end{subfigure}%
\caption{Full plots for the MNIST Wide network (zoom in for detail). 
(\subref{appfig:mnist-wide-cdf}) The CDF plot for the single-stage setting. 
(\subref{appfig:mnist-wide-all-cdf}) The CDF plot for the multi-stage setting. 
(\subref{appfig:mnist-wide-diff}) The difference plot for the single-stage setting, where the reference method is ZD-MIP. (\subref{appfig:mnist-wide-all-diff}) The difference plot for the multi-stage setting, where the reference method is AS256$\to$ZD. }\label{appfig:mnist-wide}
\end{figure}

\begin{figure}
\centering
\begin{subfigure}{0.24\textwidth}
  \centering
    \resizebox{0.99\textwidth}{!}{\input{figures/mnist_deep_cdf.pgf}} 
  \caption{Single-stage CDF}\label{appfig:mnist-deep-cdf}
\end{subfigure}%
\begin{subfigure}{0.24\textwidth}
  \centering
 \resizebox{0.99\textwidth}{!}{\input{figures/mnist_deep_all_cdf.pgf}}
  \caption{All-stage CDF}\label{appfig:mnist-deep-all-cdf}
\end{subfigure}%
\begin{subfigure}{0.24\textwidth}
  \centering
  \resizebox{0.9\textwidth}{!}{\input{figures/mnist_deep_diff.pgf}} 
  \caption{Single-stage difference}\label{appfig:mnist-deep-diff}
\end{subfigure}%
\begin{subfigure}{0.24\textwidth}
  \centering
  \resizebox{0.9\textwidth}{!}{\input{figures/mnist_deep_all_diff.pgf}}
  \caption{All-stage difference}\label{appfig:mnist-deep-all-diff}
\end{subfigure}%
\caption{Full plots for the MNIST Deep network (zoom in for detail). 
(\subref{appfig:mnist-deep-cdf}) The CDF plot for the single-stage setting. 
(\subref{appfig:mnist-deep-all-cdf}) The CDF plot for the multi-stage setting. 
(\subref{appfig:mnist-deep-diff}) The difference plot for the single-stage setting, where the reference method is ZD-MIP. (\subref{appfig:mnist-deep-all-diff}) The difference plot for the multi-stage setting, where the reference method is AS256$\to$ZD.
}\label{fig:test}\label{appfig:mnist-deep}
\end{figure}

\begin{table}[h]
\centering
\caption{Time required by each algorithm across a set of four networks in the single-stage setting.
Numbers reported are mean and standard deviations in seconds.}\label{table:single-stage-time-full}
\begin{tabular}{lrrrr}
\toprule
Alg\textbackslash{}Net & MNIST FFNet & MNIST Deep & MNIST Wide & CIFAR SGD \\
\midrule
BDD+ & \(1.7 \pm 0.1\) & \(2.5 \pm 0.2\) & \(1.5 \pm 0.1\) & \(1.5 \pm 0.1\) \\
LP & \(6.2 \pm 1.1\) & \(4.8 \pm 0.5\) & \(5.5 \pm 0.5\) & \(6.2 \pm 0.6\) \\
\textbf{ZD-2D} & \(5.7 \pm 0.3\) & \(10.9 \pm 0.5\) & \(10.2 \pm 0.5\) & \(7.6 \pm 0.3\) \\
AS & \(10.7 \pm 0.9\) & \(16.9 \pm 1.2\) & \(10.4 \pm 0.7\) & \(10.2 \pm 0.8\) \\
\textbf{ZD-MIP} & \(6.0 \pm 0.3\) & \(23.1 \pm 12.4\) & \(14.7 \pm 4.6\) & \(11.0 \pm 3.1\) \\
Anderson & \(13.1 \pm 2.5\) & \(52.4 \pm 27.5\) & \(52.4 \pm 19.2\) & \(21.3 \pm 4.6\) \\
\bottomrule
\end{tabular}
\end{table}

\begin{table}[h]
\centering
\caption{Time required by each algorithm across a set of three networks in the all-stage setting.
Numbers reported are mean and standard deviations in seconds.}\label{table:all-stage-time-full}
\begin{tabular}{lrrr}
\toprule
Alg\textbackslash{}Net & MNIST FFNet & MNIST Deep & MNIST Wide \\
\midrule
BDD+ & \(5.2 \pm 0.5\) & \(15.4 \pm 0.4\) & \(8.3 \pm 0.3\) \\
\textbf{BDD+ → ZD} & \(52.0 \pm 31.2\) & \(51.7 \pm 14.0\) & \(36.4 \pm 6.8\) \\
AS256 & \(29.5 \pm 0.9\) & \(138.9 \pm 0.9\) & \(45.4 \pm 0.6\) \\
\textbf{AS256 → ZD} & \(70.0 \pm 27.5\) & \(174.1 \pm 13.6\) & \(73.3 \pm 6.6\) \\
AS512 & \(61.6 \pm 1.8\) & \(289.3 \pm 1.7\) & \(91.3 \pm 0.9\) \\
\textbf{AS512 → ZD} & \(100.3 \pm 26.6\) & \(324.1 \pm 13.5\) & \(119.2 \pm 6.9\) \\
LP & \(90.0 \pm 16.9\) & \(2132.2 \pm 494.4\) & \(387.1 \pm 65.5\) \\
\bottomrule
\end{tabular}
\end{table}

\begin{table}[h]
\centering
\caption{Bounds obtained by each algorithm across a set of four networks in the single-stage setting. 
Numbers reported are average and standard deviation of the bound relative to \textbf{ZD-2D} over the first 1000 images from the respective validation set.}\label{table:single-stage-bound-full}
\begin{tabular}{lrrrr}
\toprule
Alg\textbackslash{}Net & MNIST FFNet & MNIST Deep & MNIST Wide & CIFAR SGD \\
\midrule
BDD+ & \(-3.4 \pm 1.1\) & \(-8.6 \pm 2.7\) & \(-4.2 \pm 1.3\) & \(-1.5 \pm 0.9\) \\
LP & \(-2.8 \pm 1.0\) & \(-6.6 \pm 2.4\) & \(-3.7 \pm 1.3\) & \(-1.2 \pm 0.9\) \\
Anderson & \(-1.3 \pm 0.7\) & \(-5.9 \pm 2.3\) & \(-2.6 \pm 1.2\) & \(-0.7 \pm 0.8\) \\
AS & \(-1.2 \pm 0.9\) & \(-6.3 \pm 2.4\) & \(-2.6 \pm 1.2\) & \(-0.5 \pm 0.8\) \\
\textbf{ZD-2D} & \(0.0 \pm 0.0\) & \(0.0 \pm 0.0\) & \(0.0 \pm 0.0\) & \(0.0 \pm 0.0\) \\
\textbf{ZD-MIP} & \(3.7 \pm 1.4\) & \(18.1 \pm 5.8\) & \(9.0 \pm 2.6\) & \(4.9 \pm 2.8\) \\
\bottomrule
\end{tabular}
\end{table}

\begin{table}[h]
\centering
\caption{Bounds obtained by each algorithm across a set of three networks in the all-stage setting. 
Numbers reported are average and standard deviation of the bound relative to \textbf{BDD+ → ZD} over the first 1000 images from the MNIST validation set.}\label{table:all-stage-bound-full}
\begin{tabular}{lrrr}
\toprule
Alg\textbackslash{}Net & MNIST FFNet & MNIST Deep & MNIST Wide \\
\midrule
BDD+ & \(-6.5 \pm 2.1\) & \(-11.5 \pm 4.5\) & \(-5.4 \pm 2.2\) \\
LP & \(-5.9 \pm 2.0\) & \(-8.2 \pm 4.0\) & \(-4.7 \pm 2.1\) \\
AS256 & \(-2.0 \pm 1.5\) & \(-5.0 \pm 3.4\) & \(-1.4 \pm 1.7\) \\
AS512 & \(-0.7 \pm 1.3\) & \(-3.1 \pm 3.1\) & \(-0.1 \pm 1.5\) \\
\textbf{BDD+ → ZD} & \(0.0 \pm 0.0\) & \(0.0 \pm 0.0\) & \(0.0 \pm 0.0\) \\
\textbf{AS256 → ZD} & \(1.1 \pm 0.4\) & \(2.3 \pm 0.7\) & \(0.9 \pm 0.8\) \\
\textbf{AS512 → ZD} & \(1.5 \pm 0.5\) & \(3.3 \pm 0.9\) & \(1.2 \pm 0.8\) \\
\bottomrule
\end{tabular}
\end{table}

\end{document}